\pdfoutput=1
\documentclass[journal]{IEEEtran}
\ifCLASSINFOpdf
\else
\fi
\usepackage{colortbl}
\usepackage{times}
\usepackage{epsfig}
\usepackage{graphicx}
\usepackage{amsmath}
\usepackage{amssymb}
\usepackage{colortbl}
\usepackage{microtype}
\usepackage{arydshln}
\usepackage{booktabs}
\usepackage{multirow}
\usepackage{pifont}
\usepackage[table]{xcolor}
\usepackage{bm}
\newcommand{\cmark}{\ding{51}}%
\newcommand{\xmark}{\ding{55}}%
\usepackage[breaklinks=true,colorlinks,bookmarks=false]{hyperref}
\begin{document}

\title{Real-time Semantic Segmentation via Spatial-detail Guided Context Propagation}

\author{Shijie Hao,
        Yuan Zhou,
        Yanrong Guo,
        Richang Hong,
        Jun Cheng, \IEEEmembership{Member,~IEEE}
        and~Meng Wang, \IEEEmembership{Fellow,~IEEE}% <-this % stops a space

\IEEEcompsocitemizethanks{\IEEEcompsocthanksitem
S. Hao, Y. Zhou, Y. Guo, R. Hong and M. Wang are with Key Laboratory of Knowledge Engineering with Big Data (Hefei University of Technology), Ministry of Education, and School of Computer Science and Information Engineering, Hefei University of Technology, Hefei 230009, China (e-mail: \href{mailto: hfut.hsj@gmail.com}{hfut.hsj@gmail.com}; \href{mailto: 2018110971@mail.hfut.edu.cn}{2018110971@mail.hfut.edu.cn}; \href{mailto: yrguo@hfut.edu.cn}{yrguo@hfut.edu.cn}; \href{mailto: hongrc.hfut@gmail.com}{ hongrc.hfut@gmail.com}; \href{mailto: eric.mengwang@gmail.com}{eric.mengwang@gmail.com}). J. Cheng is with CAS Key Laboratory of Human-Machine Intelligence-Synergy Systems, Shenzhen Institutes of Advanced Technology, Chinese Academy of Sciences, and Chinese University of Hong Kong (e-mail:\href{mailto: Jun.cheng@siat.ac.cn}{Jun.cheng@siat.ac.cn}). S. Hao is the corresponding author.}
\thanks{\qquad }}

\markboth{IEEE Transactions on Neural Networks and Learning Systems}%
{Shell \MakeLowercase{\textit{et al.}}: Bare Demo of IEEEtran.cls for IEEE Journals}

\maketitle

\begin{abstract}
Nowadays, vision-based computing tasks play an important role in various real-world applications. However, many vision computing tasks, e.g. semantic segmentation, are usually computationally expensive, posing a challenge to the computing systems that are resource-constrained but require fast response \textcolor{black}{speed}. Therefore, it is valuable to develop accurate and real-time vision processing models \textcolor{black}{that only require limited computational resources.} To this end, we propose the Spatial-detail Guided Context Propagation Network (SGCPNet) for achieving real-time semantic segmentation. In SGCPNet, we propose the strategy of spatial-detail guided context propagation. It uses the spatial details of shallow layers to guide the propagation of the low-resolution global contexts, in which the lost spatial information can be effectively reconstructed. In this way, the need for maintaining high-resolution features along the network is freed, therefore largely improving the model efficiency. On the other hand, due to the effective reconstruction of spatial details, the segmentation accuracy can be still preserved. In the experiments, we validate the effectiveness and efficiency of the proposed SGCPNet model. On the Citysacpes dataset, for example, our SGCPNet achieves $69.5 \%$ mIoU segmentation accuracy, while its speed reaches $178.5$ FPS on $768\times 1536$ images on a GeForce GTX 1080 Ti GPU card. In addition, SGCPNet is very lightweight and only contains 0.61 M parameters. The code will be released at \url{https://github.com/zhouyuan888888/SGCPNet}.
\end{abstract}


% \fi
\IEEEpeerreviewmaketitle

\section{Introduction}
\label{sec:introduction}

Vision computing plays a more and more important role in many real-world applications, e.g. human action recognition \cite{rahmani2017learning,zhang2019view}, object detection \cite{pang2020tju,li2021improving,xie2021psc} and object tracking \cite{yun2018action,zhao2015learning}, and so on. A common characteristic in these applications is that large amount of data is produced by vision sensors while a fast processing speed is required, making the system architecture solely based on cloud computing not always efficient \cite{shi2016edge}. In this context, edge computing has become a good complement, as this paradigm aims to drive computation from cloud to network edges, making the processing as close as to data generation sources.

\begin{figure}[tp!]
\centering
\includegraphics[height=3.6cm]{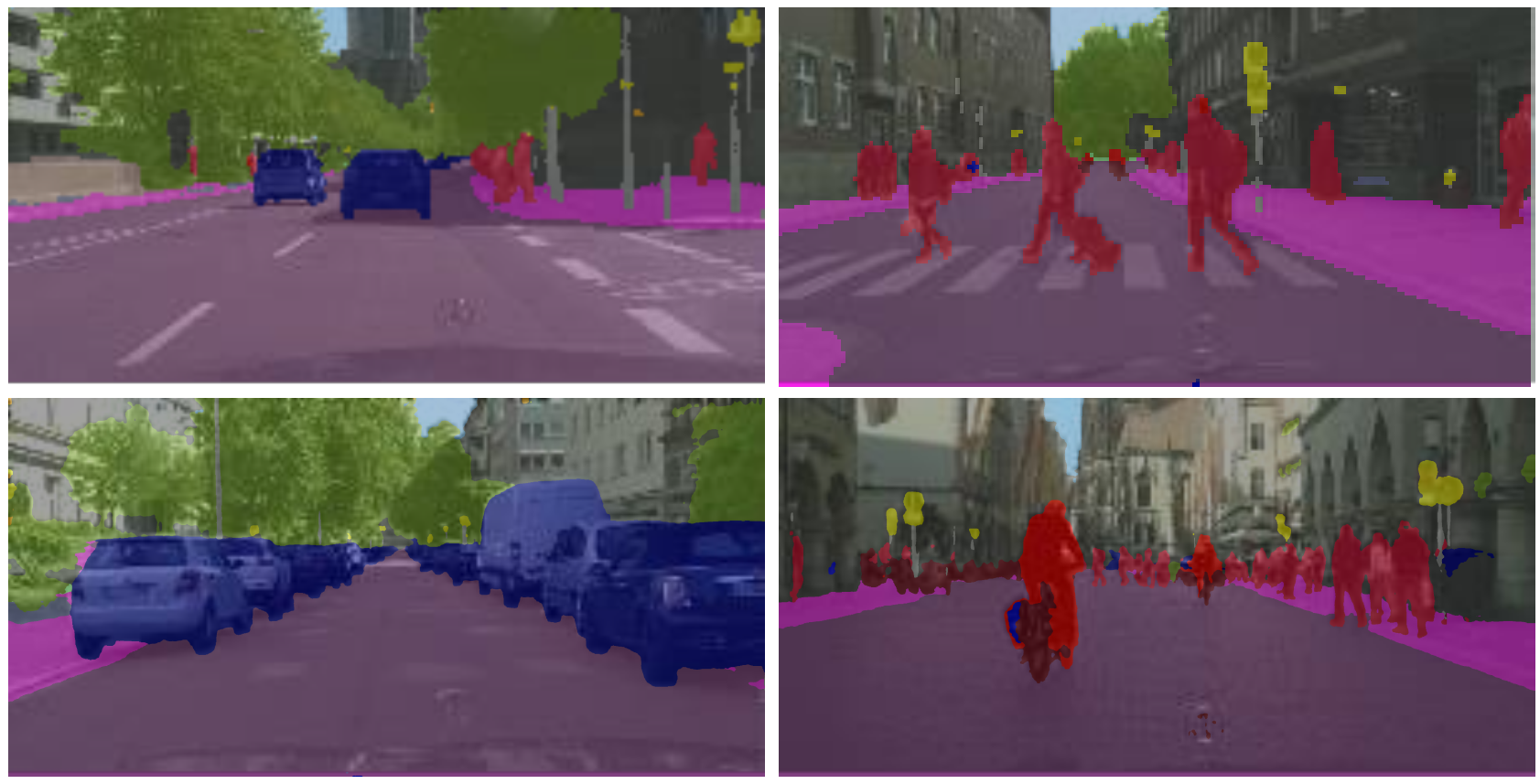}
\caption{Examples of the semantic segmentation of the road scene.}
\label{example}
\end{figure}

\begin{figure}[tp!]
\centering
\includegraphics[height=4.8cm]{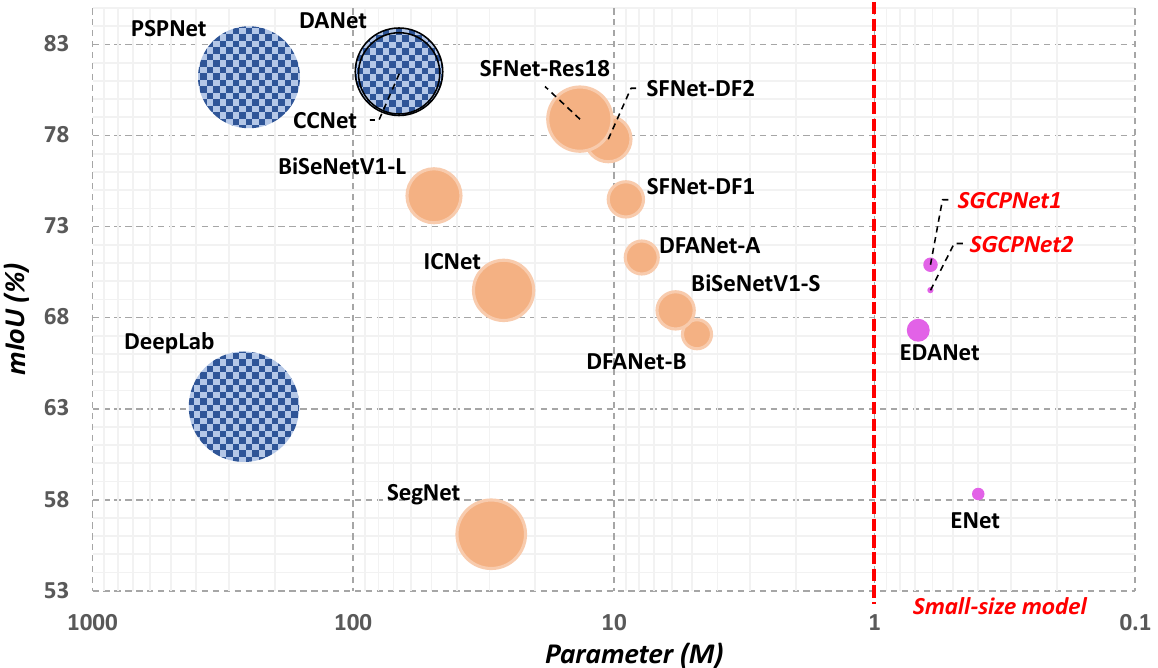}
\caption{\textcolor{black}{The balance of our SGCPNet1 and SGCPNet2 between model size, segmentation accuracy and execution speed on the Cityscapes dataset. Specially, the models of large size, medium size and small size are represented by blue, orange and pink marks, respectively.} Moreover, a smaller mark means a faster execution speed.}
\label{fig:balance_vis}
\end{figure}

Recently, the success of deep learning has simultaneously brought opportunities and challenges to edge computing \cite{PIEEE2019,wang2020convergence}. On one hand, deep learning models are able to significantly promote the system's performance. On the other hand, however, these models usually have large computation and storage costs, posing challenges to the system's implementation speed and power assumption \cite{lin2018architectural,liu2017computer}. This challenge typically exists in many vision-based computation tasks, especially for the semantic segmentation task that aims to assign each pixel with a semantic label, as shown in Fig.\ref{example}. Being able to provide pixel-level semantic information, semantic segmentation can be the cornerstone of many vision-involved applications, such as autonomous driving \cite{liu2019edge} and medical assistance systems \cite{hesamian2019deep}.

Despite the success made by the deep-learning-based semantic segmentation models  \cite{zhao2017pyramid,chen2017deeplab,fu2019dual} proposed recently, it is difficult to directly apply them in the resource-constrained scenario due to their large model size or high complexity. Model compression is a feasible way of solving these issues \cite{PIEEE2019}, such as network pruning and quantization \cite{han2015deep,bhattacharya2016sparsification,yao2017deepiot}, knowledge distillation \cite{hinton2015distilling}, and the hybrid ones \cite{liu2018demand}. On the other hand, it is more attractive to directly design a lightweight semantic segmentation model that simultaneously satisfies the following demands, i.e., being fast and accurate, while requiring low hardware consumption.

%the technique of edge-computing-based autonomous driving is a hot research direction, which aims to further improve the safety of self-driving. But this is challenging, as it requires both real-time execution speed and high decision accuracy. The self-driving system highly relies on recognition algorithms, e.g. object detection \cite{redmon2018yolov3,redmon2017yolo9000} and semantic segmentation \cite{paszke2016enet,zhao2018icnet}, which can help to realize the driving route's planning, and dynamically adjust it according to the surrounding environment.  However, since edge-computing systems of self-driving are mobile, their computation capability and memory resources are usually under strict constraints.  So, designing a memory-friendly and computation-modest recognition algorithm for autonomous driving's edge systems is critical. Moreover, its recognition performance should be accurate, which will firmly guarantee the security of users.\par

\begin{figure}[tp!]
\centering
\includegraphics[height=7.5cm]{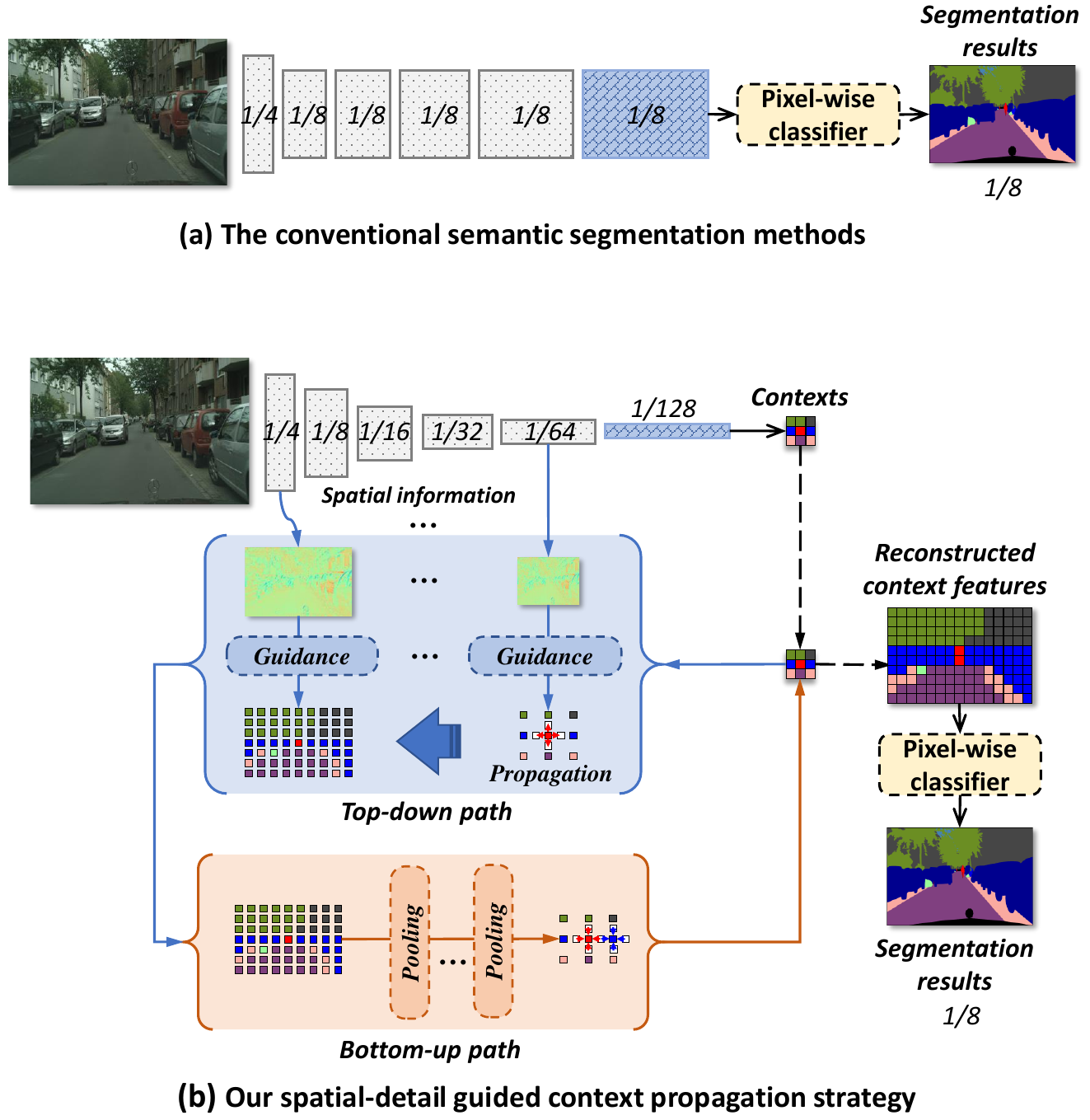}
\caption{\textcolor{black}{Schematic illustration for our spatial-detail guided context propagation strategy (b), compared with the conventional semantic segmentation methods (a) that maintain spatial details and aggregate context information  simultaneously.}}
\label{fig:context_backpropate}
\end{figure}

We first briefly analyze why the current segmentation models are computationally expensive. Aiming at high accuracy, \textcolor{black}{on one hand, the current methods \cite{liu2015parsenet,zhang2018context,he2019adaptive,chen2017deeplab,zhao2017pyramid} mainly adopt the strategy of keeping high-resolution feature maps along the network pipeline, so as to realize effective preservation of spatial detail information, as presented in Fig.\ref{fig:context_backpropate} (a). On the other hand, the increased size of feature maps relatively lowers the receptive field of convolution kernels, and thus these methods use the dilated convolution \cite{yu2015multi} to aggregate more context information as much as possible.} Nevertheless, this roadmap has a limitation that the high resolution feature maps kept in the pipeline lead to expensive computational costs. We conduct an experiment on the well-known ResNet \cite{he2016deep} to empirically verify the influence of feature resolution on Floating Point Operations (FLOPs) and runtime. Of note, for a fair comparison, the last fully-connected layer is removed. \textcolor{black}{The results in Table.\ref{table:inf_resolution} indicate that when the feature map is maintained with high resolution, the FLOPs increase substantially and the speed becomes much lower.}

Based on the above observation, we can draw that maintaining feature map resolution is the bottleneck of designing a fast and accurate segmentation model. \textcolor{black}{On one hand, it is possible to achieve a low computational complexity via downsampling the feature map quickly along the network.} The rationale is from the fact that semantic contexts belong to the region-level information, and the pixels in the neighborhood usually share a common semantic label. Therefore, it is unnecessary to always keep aggregating contexts at the high-resolution feature maps. On the other hand, however, directly using low-resolution global contexts seems to go against the demand of obtaining high-quality segmentation results, as the spatial information is severely lost in the global context features. To solve this dilemma, in this paper, our aim is to free the demand of maintaining high resolution feature maps by efficiently reconstructing the lost spatial information of the global contexts, and therefore the demands of accuracy and speed can be satisfied in the meantime.\par

%To alleviate this problem, some recent methods propose using different sub-networks to independently learn spatial details and semantic contexts \cite{zhao2018icnet,yu2018bisenet,yu2020bisenet,poudel2018contextnet}. Specifically, the network that is relatively deeper is used to learn semantic contexts, while the network with relatively shallower depth is used to maintain spatial details. However, to some extent, employing an extra network to maintain details will still hinder the improvement of the balance between accuracy and computation expenses.

\begin{table}[t]
\small
\begin{center}
\caption{\textcolor{black}{The influence of maintaining high-resolution feature maps on FLOPs and runtime. ``$\mathit{\Upsilon}$'' indicates the rate between the size of the maintained feature map and input image. Specially, in this part, the model is evaluated on an $512\times 1024$ input image.}}
\label{table:inf_resolution}
\setlength{\tabcolsep}{2.5mm}{
\begin{tabular}{c|cc|cc}
%\toprule[1.5pt]
\hline
 & \multicolumn{2}{c}{\textbf{Res-101}} & \multicolumn{2}{c}{\textbf{Res-50}} \\
 \cline{2-3}\cline{4-5}
\textbf{$\mathit{\Upsilon}$} & \textbf{FLOPs} & \textbf{Runtime} & \textbf{FLOPs} & \textbf{Runtime} \\
\hline
\hline
$1/32$ & $27.4 G$  & $37.1 ms$ &  $24.5 G$ & $24.9 ms$ \\
$1/16$ & $39.7 G$  & $85.8 ms$  & $42.1 G$  & $42.9 ms$  \\
$1/8$ & $112.5 G$ & $303.7 ms$  & $109.3 G$ &  $155.6 ms$ \\
$1/4$ & $486.3 G$ & $1207.2 ms$ & $427.2 G$ &  $542.1 ms$ \\

%\cdashline{1-4}[4pt/2pt]

%\bottomrule[1.5pt]
\hline
\end{tabular}}
\end{center}
\end{table}

%\begin{figure}[tp!]
%\centering
%\includegraphics[height=5cm]{figure/fig18.PNG}
%\caption{Comparison between high-resolution and %low-resolution context features. }
%\label{fig:comparison}
%\end{figure}

To this end, we propose the strategy of \textit{spatial-detail guided context propagation}, as shown in Fig.\ref{fig:context_backpropate} (b). It aims to use the spatial details of the shallow layers to guide the propagation of the global contexts to the neighboring positions, and therefore helps to reconstruct the lost spatial information of the global contexts. \textcolor{black}{As described in Fig.\ref{fig:context_backpropate} (b), in our method, the resolution of the feature map is reduced along the network pipeline gradually, until obtaining the final global context information.} Then, the proposed spatial-detail guided context propagation strategy is applied. We hope that the context propagation could meet two requirements. \textcolor{black}{First, during the context propagation, the context information should be consistent with its neighboring spatial details, thus guaranteeing the effectiveness of the context propagation. Second, after the propagation, the original global context information should be accurately recovered from the propagated context map as much as possible.} In this way, the propagation's accuracy can be ensured. Based on the above roadmap, we design our context propagation in a bi-direction way, i.e. 1) propagating the context information of the current position to the neighborhood, and 2) gathering the contexts from the neighboring pixels to the current position. We realize them by building the lightweight bi-directed network structure, \textcolor{black}{where we introduce the \textit{top-down path} and the \textit{bottom-up path} respectively.} Specifically, in the top-down path, the global contexts are gradually propagated to neighboring positions under the guidance of spatial details. However, in the bottom-up path, the context information of the local region is progressively gathered using the pooling operation, and thereby the global contexts can be re-extracted. The re-extracted context information is supposed to be the same as the global contexts before being propagated. \par

Based on the proposed strategy, we build a lightweight network for efficient and accurate semantic segmentation as presented in Fig.\ref{fig:SGCPNet}, named Spatial-detail Guided Context Propagation Network (SGCPNet). Our SGCPNet has low storage and computation costs. For example, it only contains $0.61 M$ parameters, \textcolor{black}{and costs only $4.5 G$ FLOPs in segmenting an $1024\times 2048$ image.} As a result, our SGCPNet has fast implementation speed. \textcolor{black}{For example, it can realize $103.7$ FPS speed on $1024\times 2048$ inputs or $731.3$ FPS speed on $360\times 480$ inputs, based on a single GTX 1080Ti GPU card. Even with an Intel Xeon Silver 4210 CPU, segmenting a $512\times 1024$ RGB image only needs $151 ms$ runtime.} In addition to the low costs and fast speed, the accuracy of our SGCPNet is still kept on a high level. For example, on the public semantic segmentation datasets Cityscapes \cite{cordts2016cityscapes} and Camvid \cite{brostow2009semantic},  SGCPNet obtains promising segmentation performances, such as $70.9 \%$ mIoU on the Cityscapes test set and $69 \%$ mIoU on the CamVid test set. \textcolor{black}{Additionally}, in Fig.\ref{fig:balance_vis}, we provide a chart that compares our SGCPNet with recent related methods in terms of model size (x-axis), segmentation accuracy (y-axis) and execution speed (mark area). We can see that our SGCPNet achieves good balance between these performance indices, \textcolor{black}{showing the advantage of our SGCPNet in the resource-constrained semantic segmentation.}\par

The contributions of this paper are summarized as the following aspects:

\begin{itemize}
\item We propose a new spatial-detail guided context propagation strategy. It effectively reconstructs the lost spatial information in the global contexts, and thus the need for maintaining \textcolor{black}{high resolution feature maps} through the network pipeline can be freed.
\item We construct the Spatial-detail Guided Context Propagation Network (SGCPNet) that realizes the spatial-detail guided context propagation with high efficiency via the bi-directed network structure.
\item \textcolor{black}{Last but not the least}, our SGCPNet presents the competitive performance on the balance between \textcolor{black}{segmentation accuracy and efficiency.}
\end{itemize}

The rest of this paper is organized as follows. \textcolor{black}{We first review the related works in Section~\ref{sec:related_work}. Then, in Section~\ref{sec:proposed_method} and Section~\ref{sec:experiments}, we respectively introduce our proposed method and the experimental results in detail. Finally, the paper is concluded in Section~\ref{sec:conclusion}.}

\section{Related Work}
\label{sec:related_work}
In this section, \textcolor{black}{we first review the relevant approaches that concentrate on boosting segmentation accuracy, and then review the methods aiming at improving segmentation efficiency.}

\subsection{\textcolor{black}{Methods for Boosting Segmentation Accuracy}}\par
\textcolor{black}{Due to the significance of obtaining sufficient context and spatial information for the semantic segmentation task, Yu et al. \cite{yu2015multi} advance the conventional convolution operation via proposing the dilated convolution.} On one hand, \textcolor{black}{the dilated convolution enlarges the receptive field of convolution through inserting ``holes'' into the convolution kernel. In this way, more context information can be aggregated.} On the other hand, some downsampling operations (e.g., pooling) for enlarging receptive field could be avoided, and \textcolor{black}{thereby the spatial details can be maintained within the network pipeline.} Based on \cite{yu2015multi}, \textcolor{black}{Chen et al. \cite{chen2017deeplab} propose the Atrous Spatial Pyramid Pooling (ASPP) module that equips the dilated convolution with the spatial pyramid structure, thus realizing context aggregation in multiple different receptive fields. Zhao et al. \cite{zhao2017pyramid} further extend ASPP to the Pyramid Pooling Module (PPM), which additionally considers the global context information. This helps the model to understand the visual scene from a more global perspective. Aiming to help the network to perceive more context information, Huang et al. \cite{huang2019ccnet} propose the Criss-Cross Network (CCNet) that aggregates the contexts lying in the criss-cross path for all pixels. While in \cite{fu2019dual}, Fu et al. propose to simultaneously aggregate the global contexts and cross-channel dependencies. With the goal of taking the advantages of dictionary learning, Zhang et al. \cite{zhang2018context} propose to build a learnable dictionary to preserve the semantic contexts of the whole training dataset. In contrary, Ma et al. \cite{ma2020preserving} propose to design a Semantics Conformity Module (SCM) to better preserve the spatial details along the network pipeline. He et al. \cite{he2019adaptive} propose to use the global-guided local affinity to guide the aggregation of pyramid context information, making the deep-learning model be more robust to the diversity of objects' size and shape. Recently, Zhang et al. \cite{zhang2020cgnet} propose to further enhance the segmentation results via explicitly considering the guidance of the edge and salient objects.} \par

\textit{\textbf{Discussion}}. Aiming to \textcolor{black}{simultaneously} obtain sufficient context information and spatial detail information, \textcolor{black}{the above methods mainly choose to maintain spatial details along the network pipeline.} Although these methods could provide accurate segmentation results, their computations are generally expensive and execution speed tends to be far from \textcolor{black}{the demands of real-time processing.} For example, DeepLab only achieves $0.25$ FPS on $512\times 1024$ images at the cost of $457.8 G$ FLOPs, while PSPNet only achieves $0.78$ FPS on $713\times 713$ images at the cost of $412.2$ G FLOPs. Therefore, they are less suitable to be applied to the applications that are resource-constrained but require fast segmentation speed.

\begin{figure*}[tp!]
\centering
\includegraphics[height=4.3cm]{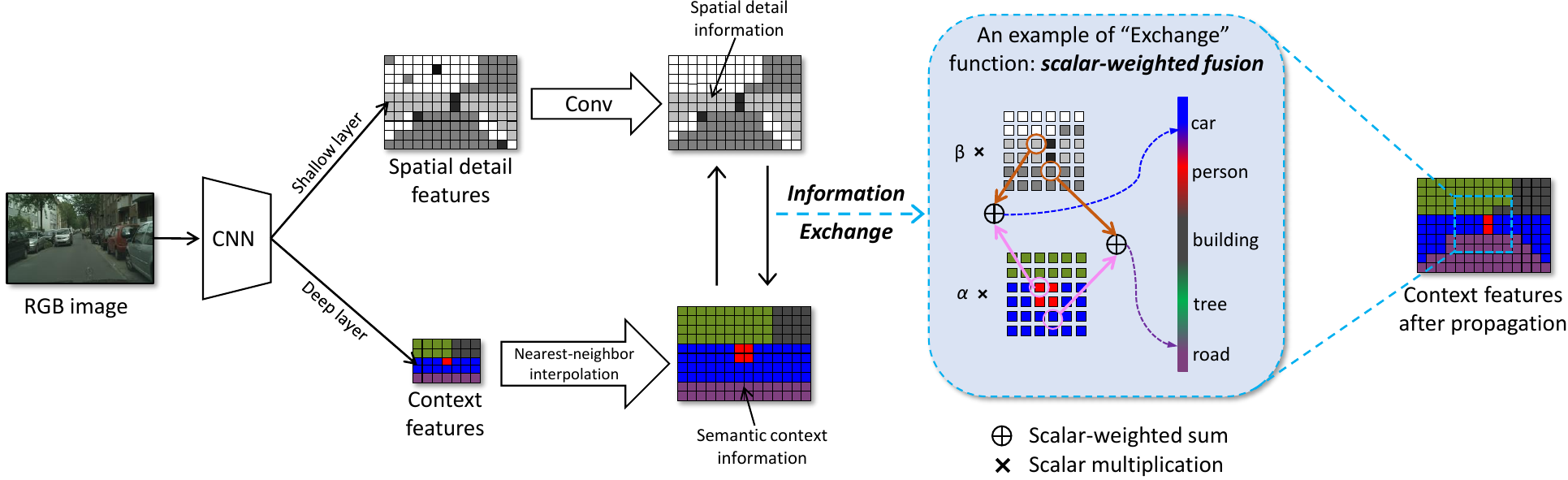}
\caption{Schematic illustration for the details of the basic operations in our spatial-detail guided context propagation.}
\label{fig:spatial_guidance}
\end{figure*}

\subsection{\textcolor{black}{Methods for Improving Segmentation Efficiency}}
In addition to \textcolor{black}{segmentation accuracy}, \textcolor{black}{segmentation efficiency} is also vital to many real-world applications. Recently, large efforts have been paid to the research of building lightweight and fast semantic segmentation models. Badrinarayanan et al. \cite{badrinarayanan2017segnet} propose an encoder-decoder model called SegNet. In the encoder, the features are gradually pooled to a low resolution, and the corresponding pooling indices are saved. In the decoder, the upsampling is performed by using the recorded pooling indices, avoiding learning how to upsample the low-resolution feature maps again, \textcolor{black}{therefore substantially improving segmentation speed.} Zhao et al. \cite{zhao2018icnet} propose ICNet that adopts multi-resolution branches, \textcolor{black}{of} which the network depths are different. \textcolor{black}{Specially,} the deep branches \textcolor{black}{are used} to extract semantic context information from the low-resolution inputs, while the shallow branches \textcolor{black}{concentrate on} capturing spatial details from the high-resolution inputs. In this way, the context information and spatial details can be both obtained, \textcolor{black}{while} the computation \textcolor{black}{costs} are saved as much as possible. \textcolor{black}{Furthermore}, Yu et al. \cite{yu2018bisenet} and Poudel et al. \cite{poudel2018contextnet} propose to separably construct a spatial path and a context path, \textcolor{black}{which are used to individually learn the spatial details and context information. Following \cite{zhao2018icnet}, in \cite{yu2018bisenet} and \cite{poudel2018contextnet}, the spatial path is kept with a shallow depth, while the context path is designed with a deep depth. In the work \cite{yu2020bisenet}, Yu et al. augment \cite{yu2018bisenet} by further introducing the guided aggregation layer that realizes better fusion between spatial details and context information. Differently, Li et al. \cite{li2019dfanet} propose the feature reuse strategy, with the goal of fully exploiting the information contained in the previous layers. Li et al. \cite{li2020semantic} propose SFNet aiming to effectively align low-resolution and high-resolution feature maps, while maintaining high efficiency. Aiming to boost segmentation's accuracy and efficiency simultaneously, Lo et al. \cite{lo2019efficient} design a network with asymmetric convolution structure and dense connection. Notably, by designing a small decoder and using the early downsampling strategy, Paszke et al. \cite{paszke2016enet} propose the lightweight ENet, which only involves about $0.4 M$ network parameters.}\par

\textit{\textbf{Discussion}}. \textcolor{black}{The current methods for improving segmentation efficiency mainly aim to individually maintain spatial details and aggregate context information, e.g. \cite{zhao2018icnet,yu2018bisenet,poudel2018contextnet,yu2020bisenet}.} We have some observations on this decoupling strategy. First, as for a typical CNN-based framework, contexts and spatial details can be obtained from the deep layers and shallow layers simultaneously. Therefore, it is possible to save some computations spend on the decoupled learning process. Second, the fusion part also needs a careful design, especially for the situations with limited computation resources. Based on these observations, \textcolor{black}{in this paper, we propose to learn spatial details and contexts from the network in the meantime. Besides, we further propose to explicitly consider the relatedness between the context and spatial detail information. We use the spatial details to guide the aggregation of context information. Of note, our SGCPNet differs from \cite{badrinarayanan2017segnet} obviously. We realize our spatial detail guidance in a learning manner, rather than using the saved pooling indices.}

%Although SegNet \cite{badrinarayanan2017segnet} uses the saved pooling indices to upsample low-resolution contexts, the information included in the pooling indices is limited.  Different from \cite{yu2018bisenet,yu2020bisenet,poudel2018contextnet}, our SGCPNet learns spatial details and context information through the same pipeline. This could free more available FLOPs and parameters, which can be more reasonably used in improving algorithms' accuracy. In particular, our SGCPNet, in which the strategy of \textit{spatial-detail guided context propagation} is used, has achieved a better balance between model size, speed and accuracy compared with the above related works, which is shown in Fig.\ref{fig:balance_vis}. This proves that our fast semantic segmentation algorithm has advantages on being applied to edge-computing-based self-driving systems.

\section{Proposed Method}
\label{sec:proposed_method}
In this section, we introduce our proposed method in detail. \textcolor{black}{We first elaborate the strategy of spatial-detail guided context propagation, and then give the details of our SGCPNet.}

\subsection{Spatial-detail Guided Context Propagation Strategy}

The motivation of \textcolor{black}{our} \textit{spatial-detail guided context propagation strategy} is simple and straightforward. \textcolor{black}{We hope the global contexts can be propagated to a higher-resolution grids, and be consistent with the low-level spatial details. To this end, the basic operations in our spatial-detail guided context propagation are built in Fig.\ref{fig:spatial_guidance}.} The context features $\bm{C}$, obtained from relatively deep layers,
are \textcolor{black}{first} upsampled to a higher resolution by the nearest neighbor interpolation operation \textit{Upsample}$(\cdot)$. \textcolor{black}{This} can be seen as a naive context propagation \textcolor{black}{which is without any help of spatial detail guidance.} Then, \textcolor{black}{we use the spatial details $\bm{S}$ of relatively shallower layers to refine the naively propagated contexts using the exchange function \textit{Exchange}$(\cdot$,$\cdot)$ that aims to facilitate the interaction between $\bm{C}$ and $\bm{S}$, therefore resulting in the reconstruction of the lost spatial information of $\bm{C}$.} The whole process can be described in Eq.\ref{eq:spatial_guide},
\begin{equation}
\bm{C}^\prime=Exchange(Upsample(\bm{C}),Conv(\bm{S}))
\label{eq:spatial_guide}
\end{equation}
where $\bm{C}^\prime$ indicates the desired contexts \textcolor{black}{propagated on a higher resolution}. \textcolor{black}{Specially, on one hand, as indicated in Eq.\ref{eq:spatial_guide},} we also adopt the convolution \textit{Conv}$(\cdot)$ to further refine \textcolor{black}{the spatial detail information of $\bm{S}$} because the features of shallow layers tend to contain \textcolor{black}{much} noise. \textcolor{black}{Considering the model efficiency, \textit{Conv}$(\cdot)$ is implemented by separable convolution \cite{chollet2017xception} in our work. On the other hand, the exchange function \textit{Exchange}$(\cdot$,$\cdot)$ can be realized in different forms.} For example, we can simply fuse them with a direct summation, or learn a pixel-wise attention mechanism. However, \textcolor{black}{aiming to facilitate a good balance between the model effectiveness and efficiency, the exchange function \textit{Exchange}$(\cdot$,$\cdot)$ is bulit} as a simple linear combination,
\begin{align}
Exchange(\bm{x_1}, \bm{x_2})=\alpha\cdot \bm{x_1}+ \beta\cdot \bm{x_2}
\label{eq:interaction}
\end{align}
where $\alpha$ and $\beta$ are learnable scalar weights for the inputs $\bm x_1$ (e.g. contexts) and $\bm x_2$ (e.g. spatial details). The learned function refines the incorrect places in the contexts by imposing a large $\beta$, \textcolor{black}{therefore} automatically \textcolor{black}{weighing} more for the spatial detail information. In the right part of Fig.\ref{fig:spatial_guidance}, we provide a toy example to explain this process. Suppose the initially upsampled $\bm{C}$ are locally inaccurate. After imposing the exchange function, these inaccurate places can be updated under the guidance of spatial detail information. For example, a few places inferred as ``person'' are corrected into the ones inferred as ``car''. Also, several places inferred as ``car'' are corrected into the ones inffered as ``road''. In this way, the upsampled contexts become more consistent with the spatial details of the original image, and \textcolor{black}{thus the lost spatial information can be recovered, as exemplified in Fig.\ref{fig:context_visuallization}.} \par

As we mentioned before, we hope that our context propagation could obey two \textcolor{black}{requirements}, i.e. 1) \textcolor{black}{during the context propagation,} the context information should be consistent with the spatial details \textcolor{black}{contained in the neighboring pixels;} 2) after \textcolor{black}{the context propagation}, the original global contexts could be recovered as much as possible. \textcolor{black}{Aiming to satisfy these two requirements,} the spatial-detail context propagation strategy is realized by building the bi-directed paths, \textcolor{black}{which we respectively call the \textit{top-down path} and the \textit{bottom-up path}. As presented in Fig.\ref{fig:context_backpropate} (b),} along the top-down path, the global contexts are back-delivered to the shallow-layers and constantly interacted with the spatial detail information, \textcolor{black}{leading to the reconstruction for the lost spatial detail information in contexts. However,} as for the bottom-up path, it aims at re-extracting the global contexts, which are supposed to be as same as the contexts before propagation. The context re-extraction can be realized by using max pooling or average pooling cooperated with convolution operations. Particularly, \textcolor{black}{in this paper}, we adopt max pooling to realize context re-extraction, \textcolor{black}{which is advantageous in selecting more discriminative responses. Also, the experimental results of ablation study presented in Table. \ref{table:ablation_TD_BP} indicate that using max pooling can yield better performance.} After the alternate usage of the top-down and bottom-up paths, the more accurate high-resolution contexts are produced, \textcolor{black}{which contains sufficient context information and spatial details simultaneously. In this way, the need for continuously maintaining high-resolution feature maps within the network can be freed, thus largely improving segmentation efficiency. Meanwhile, the segmentation accuracy is still maintained as much as possible.}

\subsection{Architecture of SGCPNet}
\begin{figure*}[tp!]
\centering
\includegraphics[height=6.5cm]{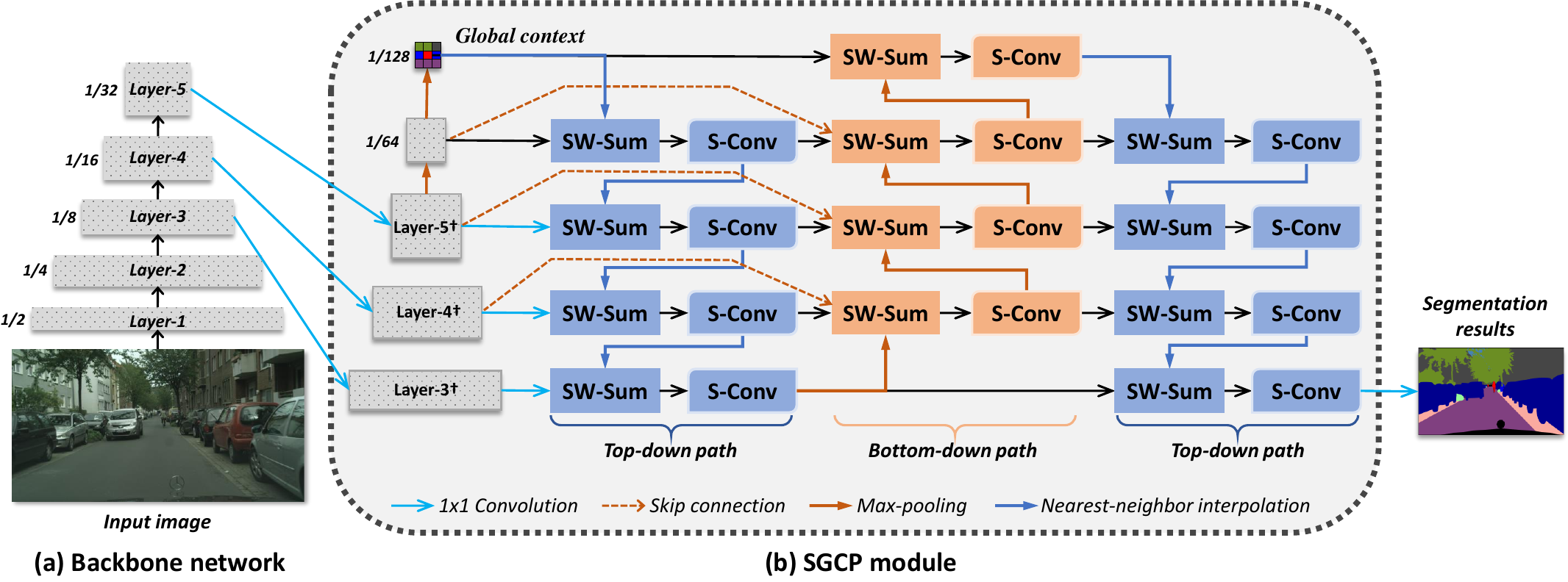}
\caption{Architecture of our Spatial-detail Guided Context Propagation Network (SGCPNet). ``SW-Sum'' represents the scalar weighted sum and ``S-Conv'' represents the separable convolution.}
\label{fig:SGCPNet}
\end{figure*}

Based on the proposed spatial-detail guided context propagation strategy, we construct the Spatial-detail Guided Context Propagation Network (SGCPNet). \textcolor{black}{As shown in Fig.\ref{fig:SGCPNet}, our} SGCPNet is a variant of encoder-decoder framework. As for the encoder, we choose the lightweight MobileNet \cite{howard2019searching} as the backbone network. \textcolor{black}{Specially, for achieving high efficiency}, we do not maintain spatial details in the \textcolor{black}{network} pipeline. \textcolor{black}{Instead, the feature map is gradually downsampled until 1/32 of the original resolution. This can substantially reduce computational expenses, and therefore endue the model with a faster segmentation speed. Nevertheless, due to the lost spatial detail information, this may lead to substantial decrease of segmentation accuracy.} To avoid this issue, we design \textcolor{black}{a} lightweight Spatial-detail Guided Context Propagation (SGCP) module as the decoder, \textcolor{black}{as shown in Fig.\ref{fig:SGCPNet} (b). It is able to effectively recover the lost spatial information of contexts, therefore segmentation accuracy can be preserved as much as possible. Aiming to enable effective context propagation at low computation consumption, the SGCP module is designed as the bi-directional structure. At last, the $1\times1$ convolutional classifier is applied to the outputs of the SGCP module, and produces the final segmentation results. In Section III-C, we provide the details of our SGCP module.}\par
%Our SGCPNet only contains 0.61 M parameters. This means it is suitable even to be applied to mobile devices, such as mobile phone. Based on the lightweight architecture, our SGCPNet can achieve 585.9 FPS based on the $360\times 640$ input images. That means that, for an image, it only costs 1.7 ms runtime. Although our SGCPNet is lightweight, it still has achieved a promising performance in accuracy. For example, it has achieved 68.4 \% mIoU  on the Cityscapes test set and 67.8 \% mIoU on the CamVid test.

\subsection{Details of SGCP module}
As shown in Fig.\ref{fig:SGCPNet} (b), three $1\times 1$ convolution operations are first applied to the features \textcolor{black}{produced by} the last three backbone layer, \textcolor{black}{so that the features are encoded into higher-dimension representations which contain more abundant feature descriptions. We term the produced three new layers as Layer-3$^{\dagger}$, Layer-4$^{\dagger}$ and Layer-5$^{\dagger}$, respectively. Then, aiming to aggregate more global context information, we successively apply two max-pooling operations to the feature map of Layer-5$^{\dagger}$.} Particularly, the kernel size of these two max-poolings is both set as $3\times 3$, and their stride is set as 2. \textcolor{black}{In this way, we obtain the final context map whose resolution is condensed to 1/128. The final context map contains more global semantic contexts, while almost all its spatial information is lost.}\par

Aiming to reconstruct the lost spatial detail information in the aggregated global contexts, the proposed spatial-detail guided context propagation strategy is used. \textcolor{black}{To realize effective context propagation, we design a bi-directed network structure, where we introduce the \textit{top-down path} and \textit{bottom-up path} respectively. Specially, the top-down path and bottom-up path have similar network structure, and they both consists of the convolution operation and the scalar-weighted fusion. Nevertheless, aiming to realize different function for these two paths, the nearest-neighbor interpolation operation is employed in the top-down path. While in the bottom-up path, we further adopt the max-pooling operation.} Following \cite{li2019dfanet,yu2020bisenet}, \textcolor{black}{we use the separable convolution \cite{chollet2017xception} to construct the convolution layers of our SGCP module, due to its modest computations and effectiveness in feature extraction.} The separable convolution is composed of a point-wise convolution and a depth-wise convolution. \textcolor{black}{Specifically, the depth-wise convolution individually extracts the features contained in each channel of the feature map, while the point-wise convolution linearly combines the features extracted by the depth-wise convolution. As presented in Fig.\ref{fig:SGCPNet}, aiming to aggregate the information coming from the neighboring layers, the scalar-weighted fusion is employed in the top-down and bottom-up paths,}
\begin{align}
\bm{F}^\prime =
\begin{cases}
\mbox{\textit{\textbf{if}} skip-connection:}\\
\qquad\alpha_{(l-1)}\cdot \bm{F}_{(l-1)}+\beta_{l}\cdot \bm{F}_{l}+\gamma_{l}\cdot \bm{S}_{l} \\
\mbox{\textit{\textbf{else}}:}\\
\qquad\alpha_{(l-1)}\cdot \bm{F}_{(l-1)}+\beta_{l}\cdot \bm{F}_{l}\\
\end{cases}
\label{eq:w-sum}
\end{align}
\textcolor{black}{where $\bm{F}_{l-1}$ and $\bm{F}_{l}$ respectively indicate the feature map coming from the $(l-1)^{th}$ and $l^{th}$ layer. Moreover, $\bm{F}_{l-1}$ contains more spatial detail information, and $\bm{F}_{l}$ contains more context information. As Eq.\ref{eq:w-sum} indicates, the formula is conditioned on whether skip connections exist. When skip connections exist, the features $\bm{S}_{l}$ coming from the $l^{th}$ layer are further considered. Additionally, the scalars $\alpha_{l-1}$, $\beta_{l}$ and $\gamma_{l}$ are learnable weights for controlling the balance between $\bm{F}_{l-1}$, $\bm{F}_{l}$ and $\bm{S}_{l}$. According to Eq.\ref{eq:w-sum},} the spatial-detail guided context propagation can be \textcolor{black}{re-interpreted} as follows: \textcolor{black}{the contexts contained in $\bm{F}_l$ are propagated to the neighborhood, via leveraging the guidance of spatial details involved in $\bm{F}_{l-1}$. Therefore, in propagated context map $\bm{F}^\prime$, the spatial detail information is reconstructed to some extent.}

\textcolor{black}{As can be seen from Fig.\ref{fig:SGCPNet} (b), in our SGCP module, we first build a top-down path, with the goal of forcing continuous interaction between the global contexts and spatial details. In this way, the global contexts are gradually propagated to the neighborhood under the guidance of spatial details, until obtaining the preliminarily reconstructed context map with the resolution at 1/8 of the input image.} Then, \textcolor{black}{we build a bottom-up path, aiming at re-extracting the global contexts from the propagated context map. Specially,} we hope that the similarity between the original global context map and the re-extracted one can be as high as possible. \textcolor{black}{Therefore, we build the skip connections as presented in Fig.\ref{fig:SGCPNet} (b), so that the original context and spatial detail information can be inserted to the corresponding layers of the bottom-up path. At last, the top-down path is constructed. It produces the final propagated context map by leveraging the information contained in the previous paths. Of note, the final pixel-wise segmentation results are produced using the $1\times 1$ convolution classifier.}  \par

\begin{figure}[tp!]
\centering
\includegraphics[height=4.5cm]{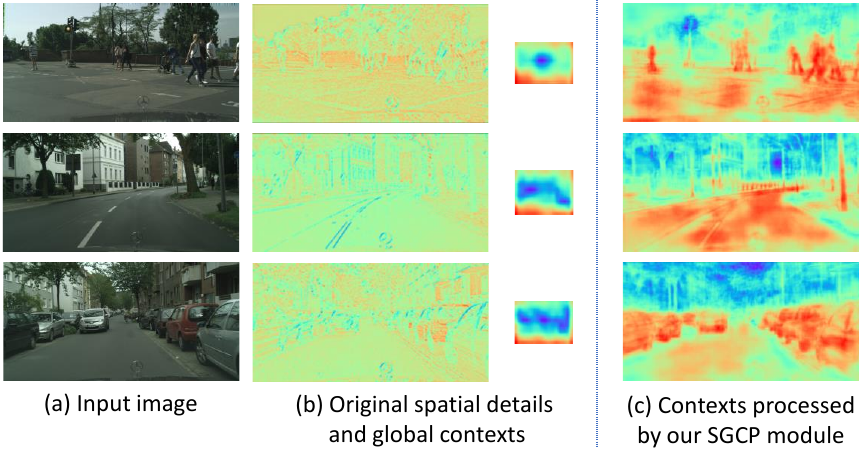}
\caption{Visualization for the context maps after being propagated under the guidance of spatial detail information by our SGCP module. As for (b) and (c), red (blue) color represents the pixel has a higher (lower) response.}
\label{fig:context_visuallization}
\end{figure}

\textcolor{black}{As indicated in Table.\ref{table:ablation_TD_BP}, our SGCP module is very lightweight, and it only contains $0.18 M$ parameters. Nevertheless, it still yields satisfying performance. For example, we visualize the global context map processed by our SGCP module in Fig. \ref{fig:context_visuallization}. As can be seen from Fig. \ref{fig:context_visuallization} (b), on one hand,} the spatial information contains much details, e.g. boundaries and textures, \textcolor{black}{but it is not aware of the scene semantics.} On the other hand, \textcolor{black}{the global context information is aware of the semantic regions at a large scale}, but its spatial details are lost severely. \textcolor{black}{As presented in Fig.\ref{fig:context_visuallization} (c), our SGCP module is able to effectively reconstruct the lost spatial information of the contexts, and thus the propagated context maps can clearly reflect the scene semantics and spatial details at the same time. Additionally, the ablation study in Section \ref{sec:ablation} also statistically validates our SGCP module's usefulness. For example, the backbone network only achieves $58.891 \%$ mIoU on the cityscapes validation set after being trained with $200$ training epochs. But based on the same training policy, the segmentation accuracy can be further improved to $68.626 \%$ mIoU by adding our SGCP module,} while only \textcolor{black}{bringing about} $0.18M$ parameters.

\begin{table*}[t]
\scriptsize
\begin{center}
\caption{Extensive comparison of parameters (Params), speed (FPS) and computation costs (FLOPs) between our method and the related works.}
\label{table:speed}
\setlength{\tabcolsep}{1.2mm}{
\begin{tabular}{r|c|cc|cc|cc|cc|cc|cc|cc|cc}
%\toprule[1.5pt]
\hline
 &  & \multicolumn{2}{c|}{$\bm{360\times480}$}& \multicolumn{2}{c|}{$\bm{360\times640}$} & \multicolumn{2}{c|}{$\bm{512\times1024}$} & \multicolumn{2}{c|}{$\bm{713\times713}$} & \multicolumn{2}{c|}{$\bm{720\times960}$} & \multicolumn{2}{c|}{$\bm{768\times1536}$}  & \multicolumn{2}{c|}{$\bm{1024\times1024}$} & \multicolumn{2}{c}{$\bm{1024\bm\times2048}$}\\
\multicolumn{1}{c|}{\textbf{Method}} & \textbf{Params} & \tiny{\textbf{FLOPs}} & \tiny{\textbf{FPS}} & \tiny{\textbf{FLOPs}} & \tiny{\textbf{FPS}} & \tiny{\textbf{FLOPs}} & \tiny{\textbf{FPS}} & \tiny{\textbf{FLOPs}} & \tiny{\textbf{FPS}} & \tiny{\textbf{FLOPs}} & \tiny{\textbf{FPS}} & \tiny{\textbf{FLOPs}} & \tiny{\textbf{FPS}} & \tiny{\textbf{FLOPs}} & \tiny{\textbf{FPS}} & \tiny{\textbf{FLOPs}} & \tiny{\textbf{FPS}} \\
%\toprule[1.5pt]
\hline
\hline
\multicolumn{1}{l|}{\textbf{Large Size}:} & & & & & & & & & & & & & & & & & \\
DeepLab \cite{chen2017deeplab} & $262.1 M$ & - & - & - & - & $457.8 G$ & $0.25$ & - & - & - & -  & - & - & - & - & - & -\\
PSPNet \cite{zhao2017pyramid} & $250.8 M$ & - & - & - & - & - & - & \cellcolor{gray!15}$412.2 G$ & \cellcolor{gray!15}$0.78$ & - & -  & - & - & - & - & - & -\\
DANet \cite{fu2019dual} & $66.6 M$ & - & - & - & - & - & - & - & - & - & - & - & - & $1111.8 G$ & $4$ & - & -\\
CCNet\cite{huang2019ccnet}  & $66.5 M$ & - & - & - & - & - & - & - & - & - & -  & - & - & $1244.8 G$ & $4.7$ & - & -\\
%\cdashline{1-18}[4pt/2pt]
\hline
\multicolumn{1}{l|}{\textbf{Medium Size}:}  & & & & & & & & & & & & & & & & &\\
SegNet \cite{badrinarayanan2017segnet} & $29.5 M$ & - & - & $286 G$ & $14.6$ & - & - & - & - & - & - & - & - & - & - & - & -\\
SQ \cite{treml2016speeding} & - & - & - & - & - & - & - & - & -  & - & -  & - & - & - & - & $270 G$ & 16.7\\
SFNet-Res18 \cite{li2020semantic} & $13.5 M$ & - & $45.2$ & - & - & - & - & - & - & - & - & - & - &  $246.5 G$ & - & - & $22$ \\
%CRF-RNN \cite{zheng2015conditional}& - & - & - & 700 & 1.4 & - & - & - & - & - & - & - & - & - & - & - & -\\
FRRN \cite{pohlen2017full} & - & - & - & - & - & $235 G$ & $2.1$ & - & - & - & - & - & - & - & - & - & - \\
FCN-8S \cite{long2015fully} & - & - & - & - & - & $136.2 G$ & $2$ & - & - & - & - & - & - & - & - & - & - \\
BiSeNetV2-L \cite{yu2020bisenet} & - & - & - & - & - & $118.5 G$ & $47.3$ & - & - & - & - & - & - & - & - & - & -  \\
%\cdashline{1-18}[4pt/2pt]
%\multicolumn{18}{l}{\textbf{Small Size}:} \\
TwoColumn \cite{wu2017real} & - & - & - & - & - & $57.2 G$ & $14.7$ & - & - & - & - & - & - & - & - & - & - \\
BiSeNetV1-L \cite{yu2018bisenet} & $49 M$ & - & - & - & $129.4$ & - & - & - & - & - & - & $55.3 G$ & $45.7$ & - & - & - & - \\
ICNet \cite{zhao2018icnet} & $26.5 M$ & - & - & - & - & - & - & - & - & - & $27.8$ & - & - & - & -  & \cellcolor{gray!15}$28.3 G$ & $30.3$ \\
SFNet-DF2 \cite{li2020semantic} & $10.5 M$ & -  & \cellcolor{gray!15}$153.8$ & - & - & - & - & - & - & - & - & - & - & $96.5 G$ & - & - & $57$ \\
SFNet-DF1 \cite{li2020semantic} & $9.0 M$ & - & - & - & - & - & - & - & - & - & - & - & - & $36.9 G$ & -  & - & \cellcolor{gray!15}$97.5$\\
BiSeNetV2-S \cite{yu2020bisenet} & - & - & - & - & - & $21.2 G$ & $156$ & - & - & - & - & - & - & -& - & - & - \\
DFANet-A \cite{li2019dfanet}& $7.8 M$ & - & - & - & - & \cellcolor{gray!15}$1.7 G$ & \cellcolor{gray!15}$160$ & - & - & - & $120$ & - & - & $3.4 G$ & $100$ & - & -\\
BiSeNetV1-S \cite{yu2018bisenet} & $5.8 M$ & - & - & - & \cellcolor{gray!15}$203.5$ & - & - & - & - & - & -  & \cellcolor{gray!15}$14.8 G$ & \cellcolor{gray!15}$72.3$ & - & - & - & - \\
DFANet-B \cite{li2019dfanet} & $4.8 M$ & - & - & - & - & - & - & - & - & - & \cellcolor{gray!15}$160$ & - & -  & \cellcolor{gray!45}$2.1 G$ & \cellcolor{gray!15}$120$ & - & - \\
%\cdashline{1-18}[4pt/2pt]
\hline
\multicolumn{1}{l|}{\textbf{Small Size}:} & & & & & & & & & & & & & & & & &\\
ContextNet \cite{poudel2018contextnet} & $0.85 M$ & - & - & - & - & - & $65.5$ & - & - & - & - & - & - & - & - & - & $18.3$ \\
EDANet \cite{lo2019efficient} & $0.68 M$ & - & - & - & - & $8.97 G$ & $81.3$ & - & - & - & - & - & - & - & - & - & - \\
ENet \cite{paszke2016enet} & \cellcolor{gray!45}$0.40 M$ & - & - & \cellcolor{gray!15}$3.8 G$ & $135.4$ & -  & - & -  & - & - & - & - & - & - & - & - & - \\
%SGCPNet-S & 0.69 M  & 2.9 & 341.3 & 6.4 & 156.3 & 11.4 & 87.7 & P & P & 14.6 & 68.4 & 27 & 37 & 13 & 77.1 & 27 & 37.1\\
%\textbf{Our SGCPNet\textcolor{red}{$\bm{\dagger}$}}& 0.61 M & 1.7 & 585.9 & 4 & 250.4 & -  & - & 5.5 & 181 & 7.4 & 134.8  & 9.8 & 102.6 & 8.6  & 116.2 & 18.2 & 55\\
\textit{\textbf{Our SGCPNet}} & \cellcolor{gray!15}$0.61 M$ & \cellcolor{gray!45}$0.38 G$ & \cellcolor{gray!45}$731.3$ & \cellcolor{gray!45}$0.51 G$ & \cellcolor{gray!45}$597.2$ & \cellcolor{gray!45}$1.13 G$ & \cellcolor{gray!45}$349.9$ & \cellcolor{gray!45}$1.12 G$ & \cellcolor{gray!45}$354.5$ & \cellcolor{gray!45}$1.49 G$ & \cellcolor{gray!45}$278.4$ & \cellcolor{gray!45}$2.53 G$ & \cellcolor{gray!45}$178.5$ & \cellcolor{gray!15}$2.25 G$ & \cellcolor{gray!45}$195$ & \cellcolor{gray!45}$4.5 G$ & \cellcolor{gray!45}$103.7$\\
%\bottomrule[1.5pt]
\hline
\end{tabular}}
\end{center}
\end{table*}

\section{Experiments}
\label{sec:experiments}
In this section, we first introduce the implementation details of our experiments, \textcolor{black}{and then compare our SPCNet with the relevant state-of-the-art approaches. At last, we conduct the ablation study to investigate the influence of each component in our approach.} \par

\subsection{Implementation Details}
All our codes are built on Pytorch\footnote{\url{https://pytorch.org}}. \textcolor{black}{We follow the previous works \cite{he2019adaptive,chen2017deeplab}, and employ the ``poly'' learning rate in the experiments, $base\_lr\times(1-\frac{iter}{total\_iter})^{power}$, where $base\_lr$ and $power$ are respectively set as $0.3$ and $0.9$.} \textcolor{black}{In addition, following \cite{zhang2020cgnet,ma2020preserving},} the data augmentation policies are adopted \textcolor{black}{to alleviate over-fitting,} e.g. we randomly flip and scale the inputs from $0.5$ to $2$. During the training phase, we choose the Stochastic Gradient Descent (SGD) optimizer, of which the momentum is set as $0.9$ and weight decay is set as $1e^{-5}$. \textcolor{black}{We validate our SGCPNet on two public semantic segmentation datasets, CamVid \cite{brostow2009semantic} and Cityscapes \cite{cordts2016cityscapes}, of which the training details are slightly different.} Specifically, as for CamVid \cite{brostow2009semantic}, \textcolor{black}{which has relatively fewer samples and lower image resolution,} we set the crop size as $720\times 720$, the batch size as $48$, and the training epoch as $250$. \textcolor{black}{Considering that Cityscapes \cite{cordts2016cityscapes} contains many high-resolution images,} we set the crop size as $1024\times1024$ and the training epoch as $350$. But due to our limited GPU resources, we set a smaller batch size 36 in this dataset. \textcolor{black}{Furthermore, in Table. \ref{table:speed}, Table. \ref{table:cityscapes} and Table. \ref{table:camvid}, we test the execution speed of our SGCPNet on a single GTX 1080Ti GPU card. While in Table. \ref{table:cpu}, our model's speed is drawn by testing on an Intel Xeon Silver 4210 CPU.}

%\begin{figure*}[tp!]
%\centering
%\includegraphics[height=6cm]{figure/fig17.PNG}
%\caption{Power consumption of our fast %semantic  segmentation algorithm.}
%\label{fig:power}
%\end{figure*}

\begin{table*}[t]
\small
\begin{center}
\caption{The results of our method on the Cityscapes test set.}
\label{table:cityscapes}
\setlength{\tabcolsep}{6mm}{
\begin{tabular}{r|ccccc}
%\toprule[1.5pt]
\hline
\textbf{Method} & \textbf{Params}& \textbf{Input Size} & \textbf{FLOPs} & \textbf{FPS} & \textbf{mIoU}\\
\hline
\hline
\multicolumn{1}{l}{\textbf{Large Size}:} & & & & &\\
DeepLab \cite{chen2017deeplab} & $262.1 M$ & $512\times1024$ & $457.8 G$ & $0.25$  & $63.1 \%$ \\
PSPNet \cite{zhao2017pyramid} & $250.8 M$ & $713\times713$ & $412.2 G$ & $0.78$  & $81.2 \%$ \\
DANet \cite{fu2019dual} & $66.6 M$ & $1024\times 1024$ & $1111.8 G$ & $4.0$ & \cellcolor{gray!45}$81.5 \%$ \\
CCNet \cite{huang2019ccnet} & $66.5 M$ & $1024\times 1024$ & $1244.8 G$ & $4.7$ & \cellcolor{gray!15}$81.4\%$ \\
%\cdashline{1-6}[4pt/2pt]
\hline
\multicolumn{1}{l}{\textbf{Medium Size}:} & & & & &\\
SegNet \cite{badrinarayanan2017segnet} & $29.5 M$ & $360\times640$ & $286.0 G$ & $14.6$ & $56.1\%$ \\
SQ \cite{treml2016speeding} & - & $1024\times2048$ & $270.0 G$ & $16.7$ & $59.8\%$ \\
SFNet-Res18 \cite{li2020semantic} & $13.5 M$ & $1024\times2048$ & $246.5 G$ & $22.0$ & $78.9\%$ \\
FRRN \cite{pohlen2017full}& - & $512\times1024$ & $235.0 G$ & $2.1$ & $71.8\%$ \\
%CRF-RNN \cite{zheng2015conditional} & - & 62.5\\
FCN-8S \cite{long2015fully} & - & $512\times1024$ & $136.2 G$ & $2.0$ & $63.1\%$ \\
BiSeNetV2-L \cite{yu2020bisenet} & - & $512\times1024$ & $118.5 G$ & $47.3$ & $75.3\%$ \\
%\cdashline{1-6}[4pt/2pt]
%\multicolumn{3}{l}{\textbf{Small Size}:} \\
TwoColumn \cite{wu2017real}& - & $512\times1024$ & $57.2 G$ & $14.7$ & $72.9\%$ \\
BiSeNetV1-L \cite{yu2018bisenet}& $49 M$ & $768\times1536$ & $55.3 G$ & $45.7$ & $74.7\%$ \\
ICNet \cite{zhao2018icnet}& $26.5 M$ & $1024\times2048$ & $28.3 G$ & $30.3$ & $69.5\%$ \\
SFNet-DF2 \cite{li2020semantic} & $10.5 M$ & $1024\times2048$ & $96.5 G$ & $57.0$ & $77.8\%$ \\
SFNet-DF1 \cite{li2020semantic} & $9.0 M$ & $1024\times2048$ & $36.9 G$ & $97.5$ & $74.5\%$ \\
BiSeNetV2-S \cite{yu2020bisenet}& - & $512\times1024$ & $21.2 G$ & \cellcolor{gray!15}$156.0$ & $72.6\%$ \\
DFANet-A \cite{li2019dfanet}& $7.8 M$ & $1024\times1024$ & $3.4 G$ & $100.0$ & $71.3\%$ \\
BiSeNetV1-S \cite{yu2018bisenet}& $5.8 M$ & $768\times1536$ & $14.8 G$ & $72.3$ & $68.4\%$ \\
DFANet-B \cite{li2019dfanet}& $4.8 M$ & $1024\times1024$ & \cellcolor{gray!45}$2.1 G$ & $120.0$ & $67.1\%$ \\
%\cdashline{1-6}[4pt/2pt]
\hline
\multicolumn{1}{l}{\textbf{Small Size}:} & & & & &\\
ContextNet \cite{poudel2018contextnet} & $0.85 M$ & $1024\times2048$ & - & $18.3$ & $66.1\%$ \\
EDANet \cite{lo2019efficient}& $0.68 M$ & $512\times1024$ & $8.97 G$ & $81.3$ & $67.3\%$ \\
ENet \cite{paszke2016enet}& \cellcolor{gray!45}$0.4 M$ & $360\times640$ & $3.8 G$ & $135.4$ & $58.3\%$ \\
%\cdashline{1-6}[4pt/2pt]\
\textit{\textbf{SGCPNet1}} & \cellcolor{gray!15}$0.61 M$ & $1024\times2048$ & $4.5 G$ & $103.7$ & $70.9\%$ \\
\textit{\textbf{SGCPNet2}} & \cellcolor{gray!15}$0.61 M$ & $768\times1536$ & \cellcolor{gray!15}$2.53 G$ & \cellcolor{gray!45} $178.5$ & $69.5\%$ \\
\hline
%\bottomrule[1.5pt]
\end{tabular}}
\end{center}
\end{table*}

\subsection{Comparison by Considering Efficiency Only}
\textcolor{black}{In this section, we compare our SGCPNet with the counterparts in terms of model efficiency, where we adopt three efficiency-related evaluation metrics,} i.e., parameter number of a model (Params), Floating Point Operations (FLOPs) and Frames Per Second (FPS).

\subsubsection{Model Categorization}
\textcolor{black}{For better comparison, the compared models are divided into three categories,} i.e., the models of (i) large size, (ii) medium size and (iii) small size, according to the following rules:
\begin{itemize}
\item The large-size model indicates the model's Params and FLOPs should be more than $50 M$ and $300 G$ simultaneously;
\item The medium-size model indicates the model's Params are between $1 M$ and $50 M$, or the FLOPs are between $10 G$ and $300 G$;
\item In the small-size model, the model's Params and FLOPs should be less than $1 M$ and $10 G$ in the meantime;
\end{itemize}
According to the above categorization, our SGCPNet typically belongs to a small-size model, as it just contains $0.61 M$ parameters \textcolor{black}{and its FLOPs are obviously less than $10 G$. For example, even segmenting a high-resolution image with $1024\times 2048$ resolution, its computational costs are only $4.5 G$ FLOPs.}\par

\subsubsection{\textcolor{black}{Performance Analysis}}
In Table.\ref{table:speed}, \textcolor{black}{we compare our SGCPNet with the related segmentation models in terms of Params, FLOPs and FPS under the different image resolutions listed in the table. Specially, for a fair comparison, we test our model's efficiency on the all  image resolutions in the table.}

\textcolor{black}{From Table.\ref{table:speed}, we can draw the following observations. Generally,} the large-size models \textcolor{black}{concentrate} on improving segmentation accuracy, \textcolor{black}{so that they tend to involve expensive computations.} For example, DeepLab \cite{chen2017deeplab} totally contains $262.1 M$ parameters. \textcolor{black}{When segmenting an $512\times 1024$ image, DeepLab needs to cost $457.8 G$ FLOPs, and its segmentation speed is only $0.25 FPS$. While PSPNet \cite{zhao2017pyramid} contains $250.8 M$ parameters, and costs $421.2 G$ FLOPs when processing an $713\times713$ input image, running at the speed of $0.78$ FPS. Despite that} CCNet \cite{huang2019ccnet} and DANet \cite{fu2019dual} have much less parameters, their FLOPs are still kept at a high level. \textcolor{black}{Additionally,} their \textcolor{black}{execution speeds} are far from the demands of real-time applications.\par

\textcolor{black}{The storage and computation burden of the medium-size models are much smaller than those of the large-size models.} Some of them achieve good performance in \textcolor{black}{model efficiency, e.g., SFNet-DF1 \cite{li2020semantic} and DFANet \cite{li2019dfanet} can realize the segmentation speed near or over $100$ FPS while their model parameters are less than 10M.}

As can be seen from Table.\ref{table:speed}, \textcolor{black}{the small-size models all have very small storage costs, and thus they are suitable to be applied to resource-restricted environments.} Our SGCPNet has the second smallest size (Params=$0.61M$) among all the listed models. It is tens or even hundreds of times smaller than the medium- or large-size competitors. Furthermore, SGCPNet has the smallest FLOPs on almost all the image resolutions. \textcolor{black}{Accordingly, our model's segmentation speed is faster than the compared models by a large margin.}

%\subsubsection{\textbf{Power Consumption of our algorithm}}
%Edge-computing-based self-driving systems are%mobile. So, their power consumption usually has strict limitations. In this part, we investigate the power consumption of our fast semantic segmentation algorithm on the GTX 1080 Ti GPU platform, based on two factors, i.e. batch size and input resolution. Batch size, a common term in machine learning, means the number of samples that are simultaneously processed by machines. So, in general, when setting a larger batch size, the machine's power consumption will correspondingly become higher. As shown in Fig.\ref{fig:power}, when we set mini-batch as 1, for $512\times 512$ inputs, our algorithm's power consumption is about 90 W. For a large input resolution $512\times 1024$, our algorithm's power consumption is also just 100 W. When we employ a larger batch size, the power consumption is still acceptable. For example, when mini-batch is 50, our algorithm just costs about 150 W power consumption on average.
\par

\subsection{Comparison by Considering the Balance between Accuracy and Efficiency}
As for the task of real-time semantic segmentation, the balance between accuracy and efficiency is a vital metric in model evaluation. Therefore, in this part, we compare our SGCPNet with the counterparts in terms of the balance between accuracy and efficiency \textcolor{black}{on the Citycapes \cite{cordts2016cityscapes} and CamVid \cite{brostow2009semantic} datasets.}

\subsubsection{Cityscapes}
\begin{figure}[tp!]
\centering
\includegraphics[height=7.6cm]{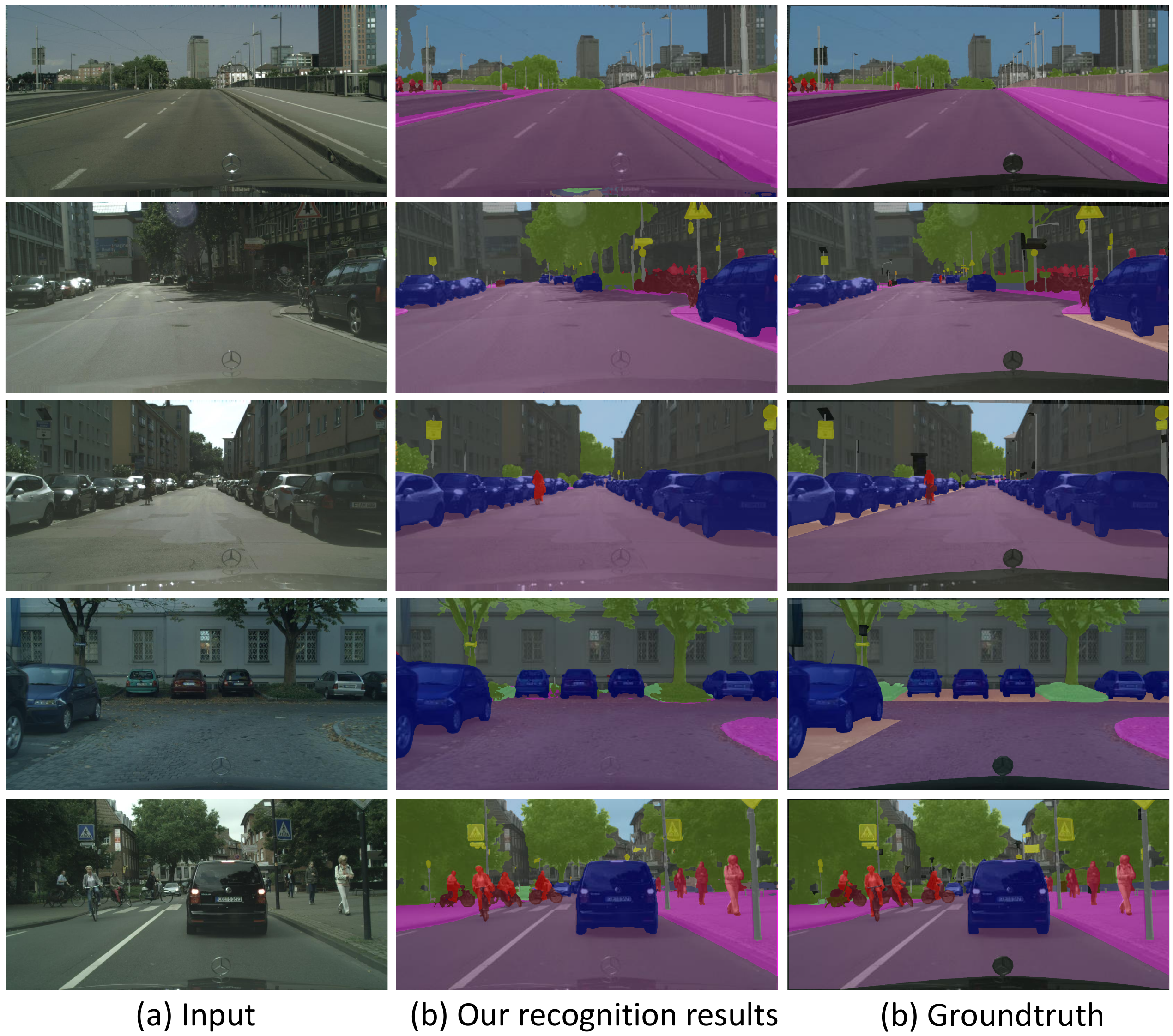}
\caption{Some typical segmentation results produced by our method on the validation set of Cityscapes.}
\label{fig:cityscapes}
\end{figure}

The Cityscapes \cite{cordts2016cityscapes} dataset contains 25000 road scene images. \textcolor{black}{Specially, $5000$ images are finely annotated, and the rest 20000 images are coarsely annotated.} In our experiments, we only employ the fine-annotated subset, \textcolor{black}{of which $2975$, $500$ and $1525$ images are respectively used for training, validating and testing.} The fine-annotated subset involves 30 semantic categories. We follow the previous works \cite{zhao2018psanet,zhao2017pyramid}, and adopt 19 categories in model evaluation. \textcolor{black}{Aiming to clearly show our model's efficiency, in this dataset, we evaluate our SGCPNet on two different image resolutions. According to different input resolutions, our model is named SGCPNet1 and SGCPNet2 respectively. As Table. \ref{table:cityscapes} indicates, as for SGCPNet1, it is evaluated on $1024\times 2048$ image resolution as same as \cite{zhao2018icnet,treml2016speeding,li2020semantic}. The high-resolution image poses more challenge for real-time semantic segmentation. While SGCPNet2 is trained and tested based on the images with a smaller $768\times 1536$ resolution.} The results of our methods and their competitors are summarized in Table.\ref{table:cityscapes}, and some examples of our segmentation results are presented in Fig.\ref{fig:cityscapes}.\par

\textit{\textcolor{black}{\textbf{Performance Analysis}}}. \textcolor{black}{Compared with the small-size models, the two versions of our SGCPNet have a better trade-off between efficiency and accuracy than that of ContextNet \cite{poudel2018contextnet} and EDANet \cite{lo2019efficient}. Although the parameters of our proposed model are slightly more than those of ENet \cite{paszke2016enet} (about $0.21 M$), our SGCPNet1 and SGCPNet2 have much higher segmentation accuracy. Additionally, our method consumes much less FLOPs. For example, even when segmenting a higher-resolution $768\times 1536$ image, our SGCPNet2 still needs much fewer FLOPs (about $1.3 G$) than those of ENet cost on processing a smaller $360\times 640$ image. Accordingly, our SGCPNet2 has much faster execution speed (about $70$ FPS).}\par

Compared with the medium-size models, our method also presents promising \textcolor{black}{trade-off between accuracy and efficiency.} For example, \textcolor{black}{on one hand,} our method has better performance than some of the models \textcolor{black}{in both efficiency and accuracy,} such as SegNet \cite{badrinarayanan2017segnet}, SQ \cite{treml2016speeding} and FCN-8S \cite{long2015fully}. \textcolor{black}{On the other hand,} the accuracy of SCGPNet1 and SGCPNet2 is at the same level of the models such as FRRN \cite{pohlen2017full}, TwoColumn \cite{wu2017real}, ICNet \cite{zhao2018icnet}, BiSeNetV2-S \cite{yu2020bisenet}, BiSeNetV1-S \cite{yu2018bisenet}, DFANet-A and DFANet-B \cite{li2019dfanet}. \textcolor{black}{But} SCGPNet1 and SGCPNet2 have much fewer FLOPs and higher FPS in most cases. \textcolor{black}{As for the models with high segmentation accuracy,} such as SFNet \cite{li2020semantic}, BiseNetV1-L \cite{yu2018bisenet} and BiseNetV2-L \cite{yu2020bisenet}, their FLOPs are tens or even one hundred times larger than ours. The comparison between these models shows that our model achieves a better balance between accuracy and efficiency, \textcolor{black}{thus more appropriate to be applied for realizing resource-constrained semantic segmentation.}

\textcolor{black}{In this part, we also compare our SGCPNet with some large-size models, aiming to prove the importance of further boosting the model efficiency. As shown in Table.\ref{table:cityscapes}, the large-size models are very advantageous in segmentation accuracy, e.g., their accuracy is around 10 \% higher than that of our SGCPNet1 and SGCPNet2.} But these methods are \textcolor{black}{extremely} not applicable in \textcolor{black}{real-time systems, due to their expensive computations and high latency. For example, segmenting an $512\times 1024$ image, DeepLab needs to cost $457.8 G$ FLOPs, and its speed is only $0.25$ FPS. In contrary, our SGCPNet has obviously higher efficiency, e.g., the FLOPs of our SGCPNet1 are only around 1/250 of DANet \cite{fu2019dual} and CCNet \cite{huang2019ccnet}.}\par

%\textcolor{red}{Compared with the medium-size models, our SGCPNet's accuracy is slightly worse. But it is still acceptable. For example, as for SQ \cite{treml2016speeding}, whose FLOPs are 60 times than ours, our method's accuracy is still higher than its by 10.6 \%. Our SGCPNet achieves 7.3 \% higher accuracy than FCN-8s \cite{long2015fully}, while we just contain its 0.8 \% FLOPs. Although the accuracies of \cite{pohlen2017full,li2020semantic,yu2020bisenet} are higher than ours, we find that they cost much more computations than our SGCPNet. For example, SFNet-ResNet18 \cite{li2020semantic} and BiSeNetV2-L \cite{yu2020bisenet} are more accurate than our SGCPNet by 8.5 \% and 4.9 \%. But the 8.5 \% and 4.9 \% accuracy improvement are respectively at the cost of additional 242 G FLOPs and 114 G FLOPs, which seems a little not cost-effective.} \par

\begin{table}[t]
\small
\begin{center}
\caption{The results of our method on the CamVid test set.}
\label{table:camvid}
\setlength{\tabcolsep}{1.5mm}{
\begin{tabular}{r|cccc}
%\toprule[1.5pt]
\hline
\multicolumn{1}{c}{\textbf{Method}} & \textbf{Params}& \textbf{Input Size} & \textbf{FPS} & \textbf{mIoU}\\
\hline
\hline
%\multicolumn{3}{l}{\textbf{Large Size}:} \\
%DPN \cite{yu2018learning} & - & 60.1\\
%DeepLab \cite{chen2017deeplab}& 262.1 M & 720$\times$960 & 4.9 & 61.6\\
%\cdashline{1-5}[4pt/2pt]
\multicolumn{2}{l}{\textbf{Medium Size}:} \\
BiSeNetV2-L \cite{yu2020bisenet} & - & $720\times960$ & $32.7$ & \cellcolor{gray!45}$73.2\%$ \\
%\cdashline{1-5}[4pt/2pt]
%\multicolumn{3}{l}{\textbf{Small Size}:} \\
BiSeNetV1-L \cite{yu2018bisenet}& $49 M$ & $720\times960$  & $116.3$ & $68.7\%$ \\
SegNet \cite{badrinarayanan2017segnet}& $29.5 M$ & $360\times480$ & $29.4$ & $55.6\%$ \\
ICNet \cite{zhao2018icnet}& $26.5 M$ & $720\times960$ & $27.8$ & $67.1\%$ \\
SFNet-Res18 \cite{li2020semantic} & $13.5 M$ & $360\times480$ & $45.2$ & \cellcolor{gray!15}$72.4\%$ \\
SFNet-DF2 \cite{li2020semantic} & $10.5 M$ & $360\times480$ & $153.8$ & $67.9\%$ \\
BiSeNetV2-S \cite{yu2020bisenet} & - & $720\times960$ & $124.5$ & \cellcolor{gray!15}$72.4\%$ \\
DFANet-A \cite{li2019dfanet}& $7.8 M$ & $720\times960$ & $120.0$ & $64.7\%$ \\
BiSeNetV1-S \cite{yu2018bisenet}& $5.8 M$ & $720\times960$ & $116.3$ & $65.6\%$ \\
DFANet-B \cite{li2019dfanet}& $4.8 M$ & $720\times960$ & \cellcolor{gray!15}$160.0$ & $59.3\%$ \\

%\cdashline{1-5}[4pt/2pt]
\hline
\multicolumn{2}{l}{\textbf{Small Size}:} \\
EDANet \cite{lo2019efficient}& $0.68 M$ & $360\times480$ & $163.0$ & $66.4\%$ \\
ENet \cite{paszke2016enet}& \cellcolor{gray!45}$0.4 M$ & $360\times480$ & $61.2$ & $51.3\%$ \\
\textit{\textbf{Our SGCPNet}} & \cellcolor{gray!15}$0.61 M$ & $720\times960$ & \cellcolor{gray!45}$278.4$ & $69.0 \%$\\
%\bottomrule[1.5pt]
\hline
\end{tabular}}
\end{center}
\end{table}

\begin{table}[t]
\small
\begin{center}
\caption{Comparison of speed and accuracy on the Cityscapes dataset when deploying the models on the CPU platform.}
\label{table:cpu}
\setlength{\tabcolsep}{1 mm}{
\begin{tabular}{r|cc|c}
%\toprule[1.5pt]
\hline
& \multicolumn{2}{c|}{\textbf{Runtime}}  & \\
\cline{2-3}
\multicolumn{1}{c|}{\textbf{Method}} & $\bm{1024\times2048}$ & $\bm{512\times1024}$ & \textbf{mIoU}\\
\hline
\hline
\multicolumn{3}{l}{\textbf{Validation set}:} &\\
MobileNet-V2-large \cite{howard2019searching} & $2980 ms$ & $786 ms$ & \cellcolor{gray!45} $72.97\%$ \\
MobileNet-V3-large \cite{howard2019searching} & $2550 ms$ & $659 ms$ & $72.37\%$ \\
MobileNet-V2-small \cite{howard2019searching} & $1270 ms$ & $354 ms$ & $66.83\%$ \\
MobileNet-V3-small \cite{howard2019searching} & \cellcolor{gray!15}{$1210 ms$} & \cellcolor{gray!15}{327 ms} & $68.38\%$ \\
\textit{\textbf{Our SGCPNet}} & \cellcolor{gray!45}{665ms} & \cellcolor{gray!45}{151ms} & \cellcolor{gray!15} $71.23\%$ \\

\cdashline{1-4}[4pt/2pt]
\multicolumn{3}{l}{\textbf{Test set}:} &\\
CCNet \cite{huang2019ccnet} & $90078 ms$ & $23060 ms$ & \cellcolor{gray!15}$81.4\%$ \\
DANet \cite{fu2019dual} & $34127 ms$ & $7380 ms$ & \cellcolor{gray!45}$81.5\%$ \\
MobileNet-V3-large \cite{howard2019searching} & $2470 ms$ & $657 ms$ & $72.6\%$ \\
MobileNet-V3-small \cite{howard2019searching} & \cellcolor{gray!15}$1030ms$ & \cellcolor{gray!15}$275ms$ & $68.3\%$ \\
\textit{\textbf{Our SGCPNet}} & \cellcolor{gray!45}$665ms$ & \cellcolor{gray!45}$151ms$ & $70.4\%$ \\

%\bottomrule[1.5pt]
\hline
\end{tabular}}
\end{center}
\end{table}

\subsubsection{CamVid}
The CamVid dataset \cite{brostow2009semantic} is collected from the high-resolution video sequences of road scenes. \textcolor{black}{This dataset totally contains $32$ semantic classes, and $701$ images. Specially, in this dataset, $367$, $101$ and $233$ images are used for training, validating, and testing respectively. Following the previous works, such as \cite{zhao2018icnet,li2019dfanet}, we only adopt $11$ classes in model evaluation.} The comparisons between the results of our method and the recent related works on this dataset are summarized
in Table.\ref{table:camvid}.\par

\textcolor{black}{\textit{\textbf{Performance Analysis}}}. Compared with \textcolor{black}{other} two small-size models, our SGCPNet achieves obviously better \textcolor{black}{trade-off between accuracy and efficiency.} For example, our SGCPNet is $115.4$ FPS faster than EDANet \cite{lo2019efficient} and $217.2$ FPS faster than ENet \cite{paszke2016enet}, \textcolor{black}{while the accuracy of our SGCPNet is still higher than that of \cite{lo2019efficient,paszke2016enet} by $2.6 \%$ and $17.7 \%$ respectively.}

\begin{table*}[t]
\small
\begin{center}
\caption{The ablation study for each component of our SGCPNet. ``BK'' denotes the backbone network and ``HFR'' indicates that the features are encoded into higher-dimension feature representations (i.e. Layer-3$\dagger$, Layer-4$\dagger$ and Layer-5$\dagger$). \textcolor{black}{``TD1'' and ``TD2'' denote the first and second top-down path,} while ``BU'' stands for the bottom-up path. Besides, ``SW-sum'' indicates the scalar-weighted sum.}
\label{table:ablation_TD_BP}
\setlength{\tabcolsep}{3.5 mm}{
\begin{tabular}{ccccccc|ccccr}
%\toprule[1.5pt]
\hline
  &  &  &  &  &  &  &  & \multicolumn{2}{c}{\textbf{Runtime}} &\\
\cline{9-10}
\textbf{BK} & \textbf{HFR} & \textbf{SW-Sum} & \textbf{TD1} & \textbf{BU} & \textbf{TD2} & \textbf{Pooling} & \textbf{Params} & \footnotesize{\textbf{GPU}} & \footnotesize{\textbf{CPU}} & \textbf{mIoU}\\
\hline
\hline
%\multicolumn{9}{l}{\textbf{W-sum}:}\\
\cmark & \xmark & \xmark  & \xmark & \xmark & \xmark & \xmark & \cellcolor{gray!45}$0.43 M$ & \cellcolor{gray!45}$1.44 ms$ & \cellcolor{gray!45}$101 ms$ & $58.891\%$ \\
%\cdashline{1-11}[4pt/2pt]
\cmark & \xmark & \cmark  & \cmark & \xmark & \xmark & $3\times3$-Max & \cellcolor{gray!15}$0.47 M$ & \cellcolor{gray!15}$1.95 ms$ & \cellcolor{gray!15}$123 ms$ & $65.292\%$ \\
\cmark & \xmark & \cmark  & \cmark & \cmark & \xmark & $3\times3$-Max & $0.50 M$ & $2.28 ms$ & $134 ms$ & $65.370\%$ \\
\cmark & \xmark & \cmark  & \cmark & \cmark & \cmark & $3\times3$-Max & $0.52 M$ & $2.74 ms$ & $153 ms$ & $67.081\%$ \\
%\cdashline{1-11}[4pt/2pt]
%\cmark & \cmark & \xmark & \xmark & \xmark & \xmark & \xmark & 0.5 M & 1.46ms & 104ms & \textcolor{red}{25.765} \\
\cmark & \cmark & \cmark & \cmark & \xmark & \xmark & $3\times3$-Max & $0.54 M$ & $2.01 ms$ & $126 ms$ & $66.345\%$ \\
\cmark & \cmark & \cmark & \cmark & \cmark & \xmark & $3\times3$-Max & $0.59 M$ & $2.40 ms$ & $135 ms$ & $67.084\%$ \\
\cmark & \cmark & \cmark & \cmark & \cmark & \cmark & $3\times3$-Max & $0.61 M$ & $2.86 ms$ & $151 ms$ & \cellcolor{gray!45}$68.626\%$ \\
\cdashline{1-11}[4pt/2pt]
\cmark & \cmark & \cmark & \cmark & \cmark & \cmark & $5\times5$-Max & $0.61 M$ & $2.75 ms$ & $155 ms$ & \cellcolor{gray!15}$68.531\%$ \\
\cmark & \cmark & \xmark & \cmark & \cmark & \cmark & $3\times3$-Max & $0.61 M$ & $2.47 ms$ & $155 ms$ & $67.792\%$ \\
\cmark & \cmark & \cmark & \cmark & \cmark & \cmark & $3\times3$-Avg & $0.61 M$ & $2.76 ms$ & $156 ms$ & $68.079\%$ \\

%\multicolumn{9}{l}{\textbf{Sum}:}\\
%\bottomrule[1.5pt]
\hline
\end{tabular}}
\end{center}
\end{table*}

\begin{figure*}[tp!]
\centering
\includegraphics[height=8cm]{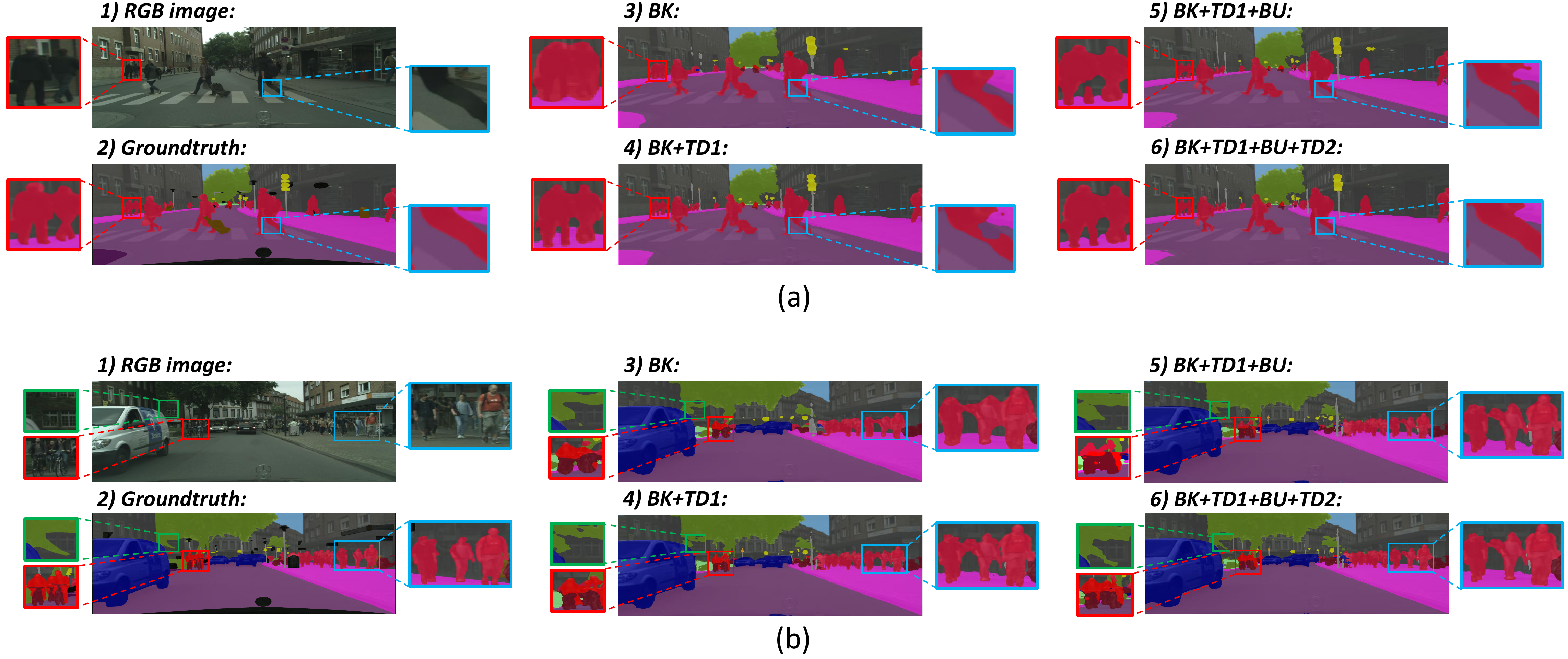}
\caption{Visualization for the influence of our top-down and bottom-up path for the final segmentation results.}
\label{fig:tb}
\end{figure*}

\textcolor{black}{Compared with} the medium-size models, the accuracy of our SGCPNet still ranks the fourth place, while it is much smaller and faster than all the medium-size models. For example, the model size of our SGCPNet is twenty times smaller than that of SFNet-Res18 \cite{li2020semantic}. \textcolor{black}{In the meantime}, our SGCPNet is several times faster than the two versions of BiSeNetV2 \cite{yu2020bisenet}. As for the rest seven medium-size models, SGCPNet consistently outperforms them in both accuracy and efficiency. \textcolor{black}{The experimental results on Camvid indicate that our SGCPNet is competent for the resource-constrained situations.}

\subsection{Ablation Study}
\label{sec:ablation}

In this part, we conduct ablation studies for our method. \textcolor{black}{We first show the performance of our model on the CPU hardware platform, and then investigate the influence of each component of our SGCPNet.}\par

\subsubsection{Performance on Different Hardware Platforms}
\textcolor{black}{Considering that the speed reported in Table.\ref{table:speed}, Table.\ref{table:cityscapes} and Table.\ref{table:camvid} is evaluated on the GPU platform, we further investigate our model's execution speed on the CPU platform that is equipped with an Intel Xeon Silver 4210 CPU.} As can be seen from Table.\ref{table:cpu}, our method still yields promising performance. Specifically, \textcolor{black}{for segmenting an $512\times 1024$ input image,} our SGCPNet only needs $151 ms$ runtime. E\textcolor{black}{ven for a large resolution $1024\times 2048$, SGCPNet still just costs $665 ms$ runtime in segmenting an input image.} These results show that our model also achieves relatively better efficiency on the CPU platform. Besides, from Table.\ref{table:cpu}, we can observe that our model has better performance than another lightweight model MobileNet-V3-small in terms of both segmentation accuracy and speed.

\subsubsection{Ablation Study on the Components in SGCPNet}
In this part, we investigate the contribution of each component in SGCPNet in terms of accuracy, speed and model size. Without losing generality, we only adopt 200 training epochs in experiments for convenience. The results are summarized in Table.\ref{table:ablation_TD_BP}.\par
From Table.\ref{table:ablation_TD_BP}, we can see that the backbone network (BK), containing $0.43 M$ parameters, achieves $58.891 \%$ mIoU on the Cityscapes validation set. \textcolor{black}{By adding the first top-down path (TD1)}, the accuracy increases to $65.292 \%$ mIoU, at the cost of additional $0.04 M$ parameters and $0.51 ms$ GPU runtime. \textcolor{black}{After being further equipped with a bottom-up path (BU), the model achieves $65.370 \%$ mIoU segmentation accuracy.} Correspondingly, the GPU runtime and parameters increase to $2.28 ms$ and $0.5 M$, respectively. When the last top-down path \textcolor{black}{(TD2)} is employed, the model performance reaches $67.081 \%$ mIoU, at the cost of $0.02 M$ parameters and $0.46 ms$ GPU runtime. \textcolor{black}{In addition, we also visualize some typical segmentation results in Fig.\ref{fig:tb}. From the figure, the backbone network (BK) generally fails in clearly segmenting the contours of objects. However, by using our top-down and bottom-up paths, the segmentation results become more and more accurate. These experimental results again empirically validate the effectiveness and efficiency of our bi-directional paths.} \par
When we further introduce the high-dimension feature representations (HFR) into the model, the accuracy is further improved to $68.626 \%$ mIoU, at the cost of only $0.09 M$ parameters and $0.12 ms$ GPU runtime.
We further investigate the influence of scalar-weighted sum (SW-Sum) and different types of pooling operations (Pooling) used in the SGCP module. When the scalar-weighted sum is replaced with the conventional sum operation, the accuracy has the decrease of nearly $1 \%$ mIoU \textcolor{black}{because of treating input feature maps equally}. \textcolor{black}{As can be seen from Table.\ref{table:ablation_TD_BP}, replacing $3\times 3$ max pooling with $3\times 3$ average pooling also leads to the decrease of accuracy. Furthermore, by enlarging the kernels of the max pooling to $5\times 5$, the performance decreases to $68.531 \%$ mIoU as well.}\par

\textcolor{black}{According to the results presented in Table.\ref{table:ablation_TD_BP}, on one hand, we can conclude that each component in our SGCPNet contributes to the final segmentation accuracy. On the other hand, the increased model size and runtime brought by them (both on GPU and CPU) are small and acceptable.}

\section{Conclusion}
\label{sec:conclusion}
In this paper, we \textcolor{black}{design} the Spatial-detail Guided Context Propagation Network (SGCPNet) for the real-time semantic segmentation task. SGCPNet frees the need of maintaining the high-resolution feature map in the network pipeline, while still well incorporating the context and spatial detail information. Therefore, SGCPNet is able to achieve the state-of-the-art balance on segmentation accuracy and efficiency, which is very suitable for being applied to the resource-constrained \textcolor{black}{systems}. \textcolor{black}{Extensive experiments on the CamVid and Cityscapes datasets demonstrate the effectiveness and efficiency of our method. For example, our SGCPNet achieves $70.9 \%$ mIoU on the Cityscapes dataset, with $0.61 M$ model size and over $100$ FPS execution speed on $1024 \times 2048$ input images.} \textcolor{black}{In the future, we plan to advance our SGCPNet to realize real-time semantic segmentation in few shots scenario, thereby relieving the intensive labor spent on data collection and annotation, and further boosting the efficiency of the semantic segmentation technique in practical applications.}

\section*{Acknowledgement}
This work was supported by National Key Research and Development Program under Grant No. 2019YFA0706200, the National
Nature Science Foundation of China under Grant No. 62172137, 62072152, 61725203 and the Fundamental Research Funds for the Central Universities under Grant No. PA2020GDKC0023.

% or


%\appendices
%% you can choose not to have a title for an appendix
%% if you want by leaving the argument blank
%\section{}
%% use section* for acknowledgment
%\section*{Acknowledgment}

\ifCLASSOPTIONcaptionsoff
  \newpage
\fi

\bibliographystyle{IEEEtran}
\bibliography{ref}

\begin{IEEEbiography}[{\includegraphics[width=1in,height=1.25in,clip,keepaspectratio]{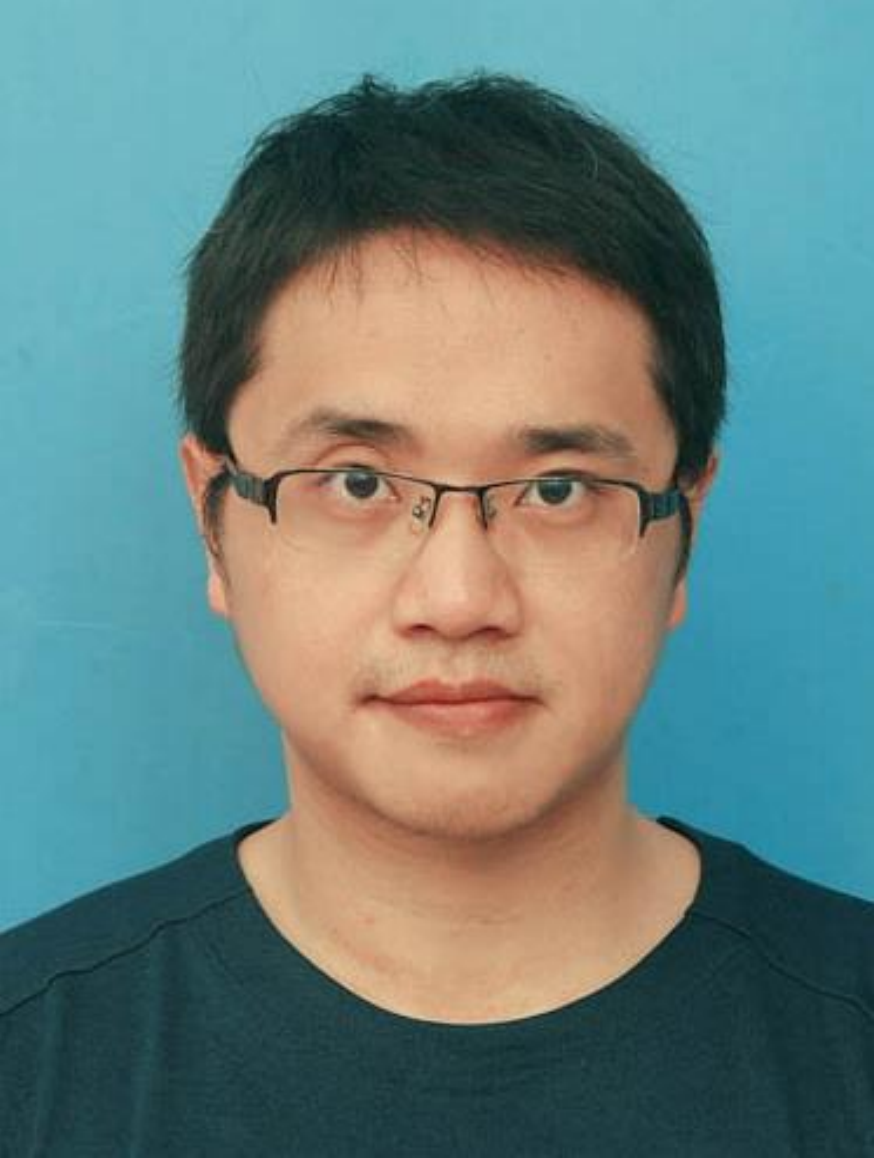}}]{Shijie Hao}
is an associate professor at School of Computer Science and Information Engineering, Hefei University of Technology (HFUT). He is also with Key Laboratory of Knowledge Engineering with Big Data (Hefei University of technology), Ministry of Education. He received his Ph.D. degree at HFUT in 2012. His research interests include image processing and multimedia content analysis.
\end{IEEEbiography}

\begin{IEEEbiography}[{\includegraphics[width=1in,height=1.25in,clip,keepaspectratio]{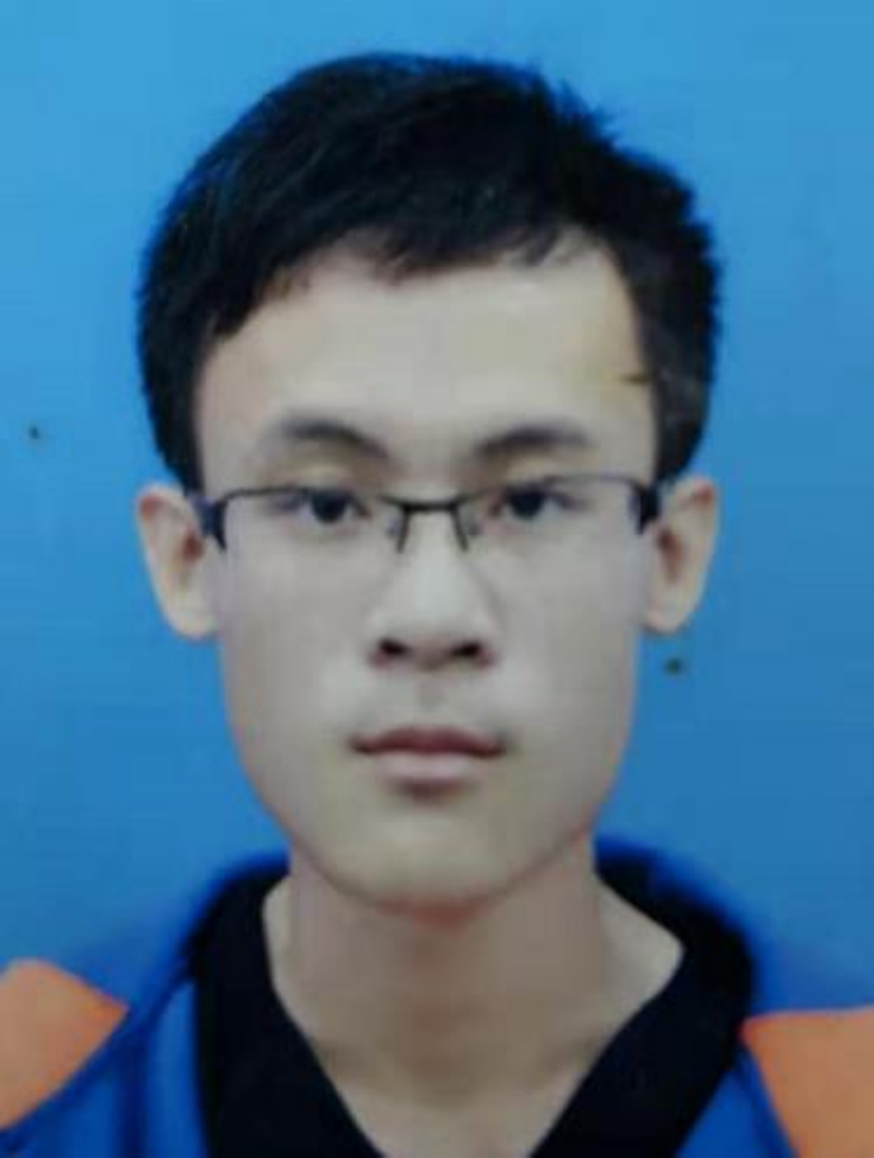}}]{Yuan Zhou}
is pursuing his Ph.D Degree at School of Computer and Information, Hefei University of Technology. He is also with Key Laboratory of Knowledge Engineering with Big Data (Hefei University of technology), Ministry of Education. His research interests include image segmentation and few-shot learning.
\end{IEEEbiography}

\begin{IEEEbiography}[{\includegraphics[width=1in,height=1.25in,clip,keepaspectratio]{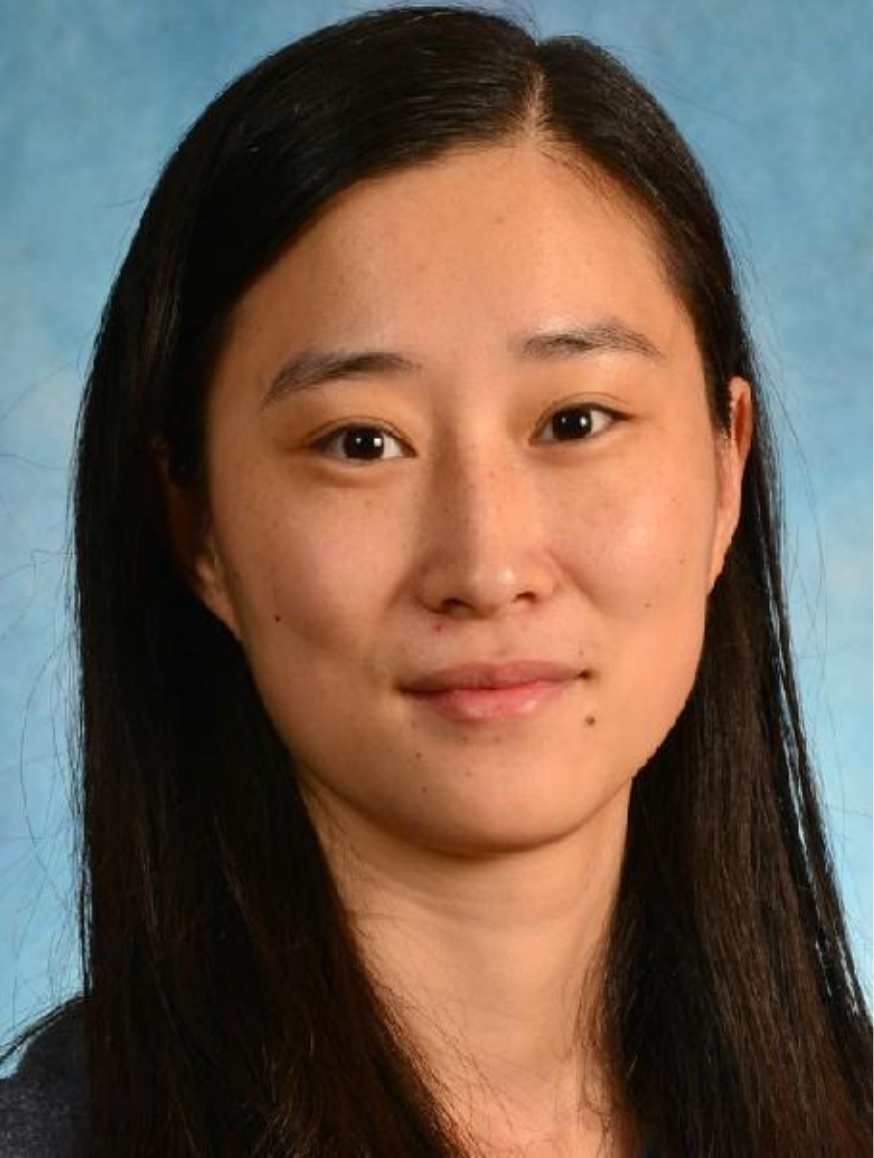}}]{Yanrong Guo}
is an associate professor at School of Computer and Information, Hefei University of Technology (HFUT). She is also with He is also with Key Laboratory of Knowledge Engineering with Big Data (Hefei University of technology), Ministry of Education. She received her Ph.D. degree at HFUT in 2013. She was a postdoc researcher at University of North Carolina at Chapel Hill (UNC) from 2013 to 2016. Her research interests include biomedical image segmentation and analysis.
\end{IEEEbiography}

\begin{IEEEbiography}[{\includegraphics[width=1in,height=1.25in,clip,keepaspectratio]{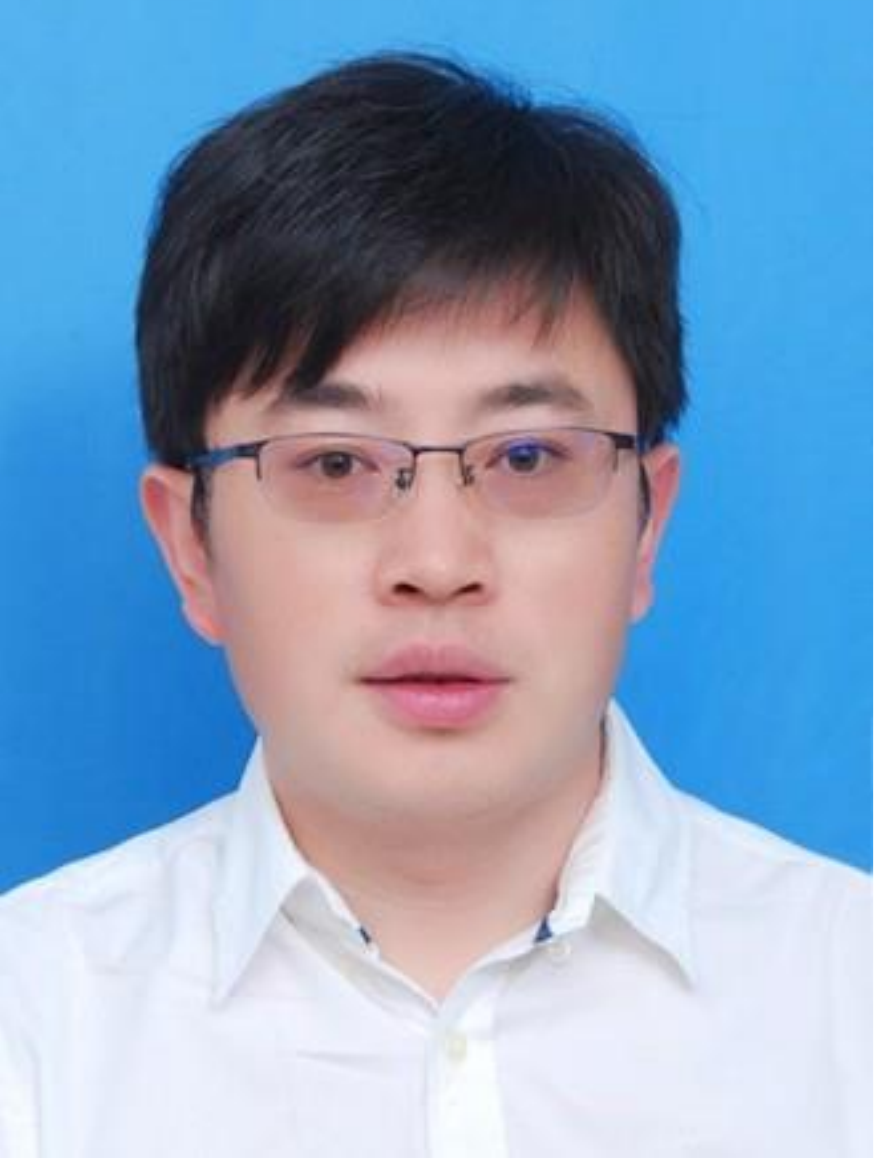}}]{Richang Hong}
received the Ph.D. degree from the University of Science and Technology of China, Hefei, China, in 2008. He was a Research Fellow of the School of Computing with the National University of Singapore, from 2008 to 2010. He is currently a Professor with the Hefei University of Technology, Hefei. He is also with Key Laboratory of Knowledge Engineering with Big Data (Hefei University of technology), Ministry of Education. He has coauthored over 70 publications in the areas of his research interests, which include multimedia content analysis and social media. He is a member of the ACM and the Executive Committee Member of the ACM SIGMM China Chapter. He was a recipient of the Best Paper Award from the ACM Multimedia 2010, the Best Paper Award from the ACM ICMR 2015, and the Honorable Mention of the IEEE Transactions on Multimedia Best Paper Award. He has served as the Technical Program Chair of the MMM 2016. He has served as an Associate Editor of IEEE Multimedia Magazine, Neural Processing Letter (Springer) Information Sciences (Elsevier) and Signal Processing (Elsevier).
\end{IEEEbiography}

\begin{IEEEbiography}[{\includegraphics[width=1in,height=1.25in,clip,keepaspectratio]{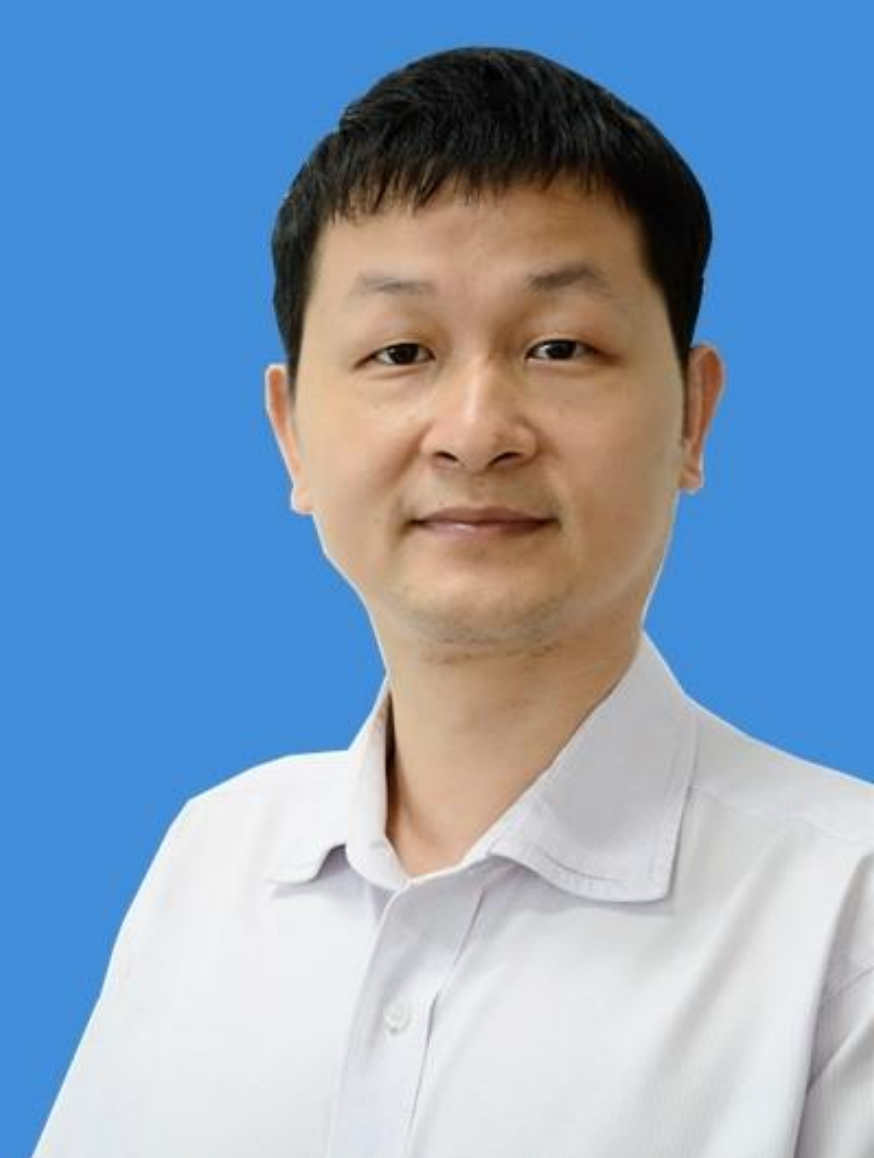}}]{Jun Cheng}
received the B.Eng. and M.Eng. degrees from the University of Science and Technology of China, Hefei, China, in 1999 and 2002, respectively, and the Ph.D. degree from the Chinese University of Hong Kong, Hong Kong, in 2006. He is currently a Professor with the Shenzhen Institutes of Advanced Technology, Chinese Academy of Sciences, Shenzhen, China, and the Director of the Laboratory for Human Machine Control. His current research interests include computer vision, robotics, machine intelligence, and control.
\end{IEEEbiography}

\begin{IEEEbiography}[{\includegraphics[width=1in,height=1.25in,clip,keepaspectratio]{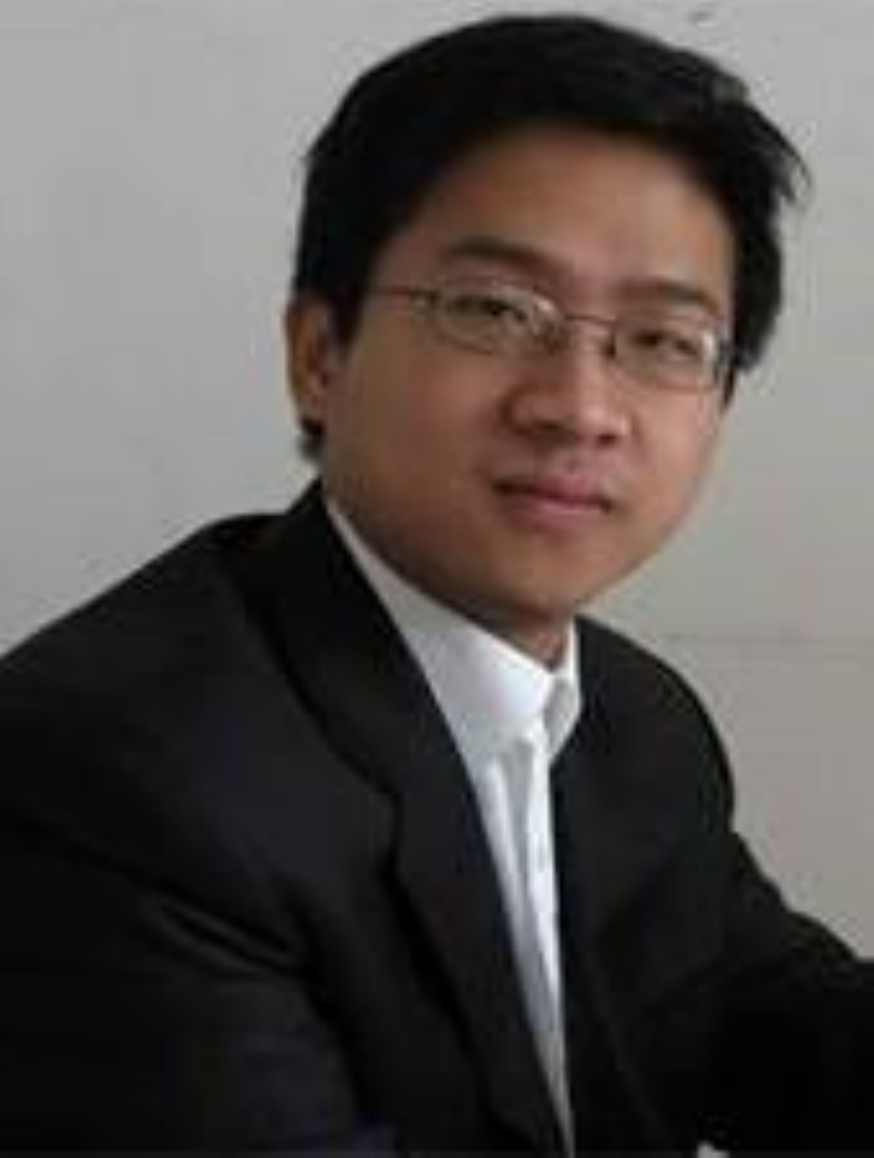}}]{Meng Wang}
received the BE and PhD degrees in the special class for the Gifted Young and the Department of Electronic Engineering and Information Science from the University of Science and Technology of China (USTC), Hefei, China, in 2003 and 2008, respectively. He is a professor at the Hefei University of Technology, China. He is also with Key Laboratory of Knowledge Engineering with Big Data (Hefei University of technology), Ministry of Education. His current research interests include multimedia content analysis, computer vision, and pattern recognition. He has authored more than 200 book chapters, journal and conference papers in these areas. He is the recipient of the ACM SIGMM Rising Star Award 2014. He is an associate editor of the IEEE Transactions on Knowledge and Data Engineering (IEEE TKDE), the IEEE Transactions on Circuits and Systems for Video Technology (IEEE TCSVT), and the IEEE Transactions on Neural Networks and Learning Systems (IEEE TNNLS).
\end{IEEEbiography}

\end{document}